\documentclass[twoside,preprint]{article}
\usepackage{ecj,palatino,epsfig,latexsym,natbib}

\usepackage{booktabs} % For formal tables
\usepackage{enumitem}
\usepackage[ruled, noend]{algorithm2e}
\usepackage{url}
\usepackage{times} % assumes new font selection scheme installed
\usepackage{lipsum}  
\usepackage{graphicx}
\usepackage{amsmath}
\usepackage{wrapfig}
\usepackage{subfiles}
\usepackage{amsfonts}
\usepackage[acronym]{glossaries}
\usepackage{hyperref}
\usepackage{textcomp}
\usepackage{gensymb}
\usepackage{nicefrac}
\usepackage{float}

\newacronym{bs}{$\mathcal{B}$}{behavior space}
\newacronym{ae}{AE}{autoencoder}
\newacronym{rl}{RL}{Reinforcement Learning}
\newacronym{ns}{NS}{Novelty Search}
\newacronym{qd}{QD}{Quality-Diversity}
\newacronym{dmp}{DMP}{Dynamic Movement Primitive}
\newacronym{10d}{10D}{10-dimensional}
\newacronym{cnn}{CNN}{Convolutional Neural Network}
\newacronym{bc}{BC}{Behaviour characterization}
\newacronym{nn}{NN}{Neural Networks}
\newacronym{dof}{DoF}{Degrees of Freedom}
\newacronym{es}{ES}{Evolutionary Strategy}
\newacronym{ea}{EA}{Evolutionary Algorithm}
\newacronym{me}{ME}{MAP-Elites}
\newacronym{serene}{SERENE}{SparsE Reward Exploration via Novelty search and Emitters}
\newacronym{taxons}{TAXONS}{Task Agnostic eXploration of Outcome spaces through Novelty and Surprise}
\newacronym{name}{STAX}{SERENE augmented TAXONS}
\newacronym{moo}{MOO}{Multi-Objective optimization}
\newacronym{im}{IM}{Intrinsic Motivation}
\newacronym{gep}{GEP}{Goal-Exploration Processes}
\newacronym{rnd}{RND}{Random Network Distillation}

\newcommand\SLASH{\char`\\}

\usepackage{xcolor}
\usepackage[draft]{changes}  % simply set the "final" option when you want to get rid of the markup
\definecolor{color-alban}{rgb}{0.0, 0.4, 0.9}
\definechangesauthor[name={Alban}, color=color-alban]{A}

\definecolor{color-giuseppe}{rgb}{1, 0.0, 0.0}
\definechangesauthor[name={Giuseppe}, color=color-giuseppe]{G}

\definecolor{color-stephane}{rgb}{0.4, 0.0, 0.9}
\definechangesauthor[name={Stephane}, color=color-stephane]{S}

\definecolor{color-alex}{rgb}{0.5, 0.8, 0.0}
\definechangesauthor[name={Alex}, color=color-alex]{AC}

\begin{document}
\ecjHeader{x}{x}{xxx-xxx}{201X}{Discovering and Exploiting Sparse Rewards in a Learned Behavior Space}{G. Paolo, M. Coninx, A. Laflaquiere, and S. Doncieux}
\title{\bf Discovering and Exploiting Sparse Rewards in a Learned Behavior Space}  

\author{\name{\bf Giuseppe Paolo} \hfill \addr{giuseppe.paolo@softbankrobotics.com}\\
       \addr{AI Lab, SoftBank Robotics Europe\\ 
             Sorbonne Universit\'{e}, CNRS, Institut des Syst\`{e}mes Intelligents et de Robotique, ISIR\\
             Paris, France}
       \AND
       \name{\bf Miranda Coninx} \hfill \addr{miranda.coninx@sorbonne-universite.fr}\\
       \addr{Sorbonne Universit\'{e}, CNRS, Institut des Syst\`{e}mes Intelligents et de Robotique, ISIR\\
       Paris, France}
       \AND
       \name{\bf Alban Laflaqui\`{e}re} \hfill \addr{alaflaquiere@softbankrobotics.com}\\
       \addr{AI Lab, SoftBank Robotics Europe\\ Paris, France}
       \AND
       \name{\bf Stephane Doncieux} \hfill \addr{stephane.doncieux@sorbonne-universite.fr}\\
       \addr{Sorbonne Universit\'{e}, CNRS, Institut des Syst\`{e}mes Intelligents et de Robotique, ISIR\\
       Paris, France}
}
\maketitle

\begin{abstract}
\looseness=-1
Learning optimal policies in sparse rewards settings is difficult as the learning agent has little to no feedback on the quality of its actions. 
In these situations, a good strategy is to focus on exploration, hopefully leading to the discovery of a reward signal to improve on. 
A learning algorithm capable of dealing with this kind of settings has to be able to (1) explore possible agent behaviors and (2) exploit any possible discovered reward. 
Exploration algorithms have been proposed that require the definition of a low-dimension behavior space, in which the behavior generated by the agent's policy can be represented.
The need to design a priori this space such that it is worth exploring is a major limitation of these algorithms. In this work, we introduce STAX, an algorithm designed to learn a behavior space on-the-fly and to explore it while optimizing any reward discovered. 
It does so by separating the exploration and learning of the behavior space from the exploitation of the reward through an alternating two-step process. 
In the first step, STAX builds a repertoire of diverse policies while learning a low-dimensional representation of the high-dimensional observations generated during the policies evaluation. 
In the exploitation step, emitters optimize the performance of the discovered rewarding solutions. 
Experiments conducted on three different sparse reward environments show that STAX performs comparably to existing baselines while requiring much less prior information about the task as it autonomously builds the behavior space it explores.
\end{abstract}


\section{Introduction}
\label{sec:intro}
\setlength{\textfloatsep}{0.25cm}
\begin{figure}[t]
    \centering
    \includegraphics[width=.85\linewidth]{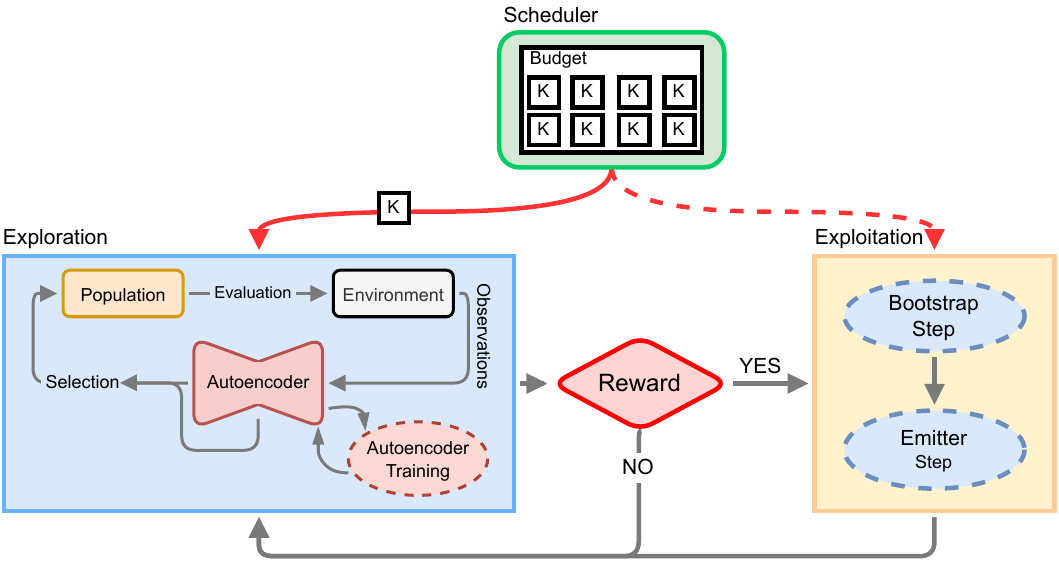}
    \vspace*{-4mm}
    \caption{STAX consists of an exploration and an exploitation process alternating thanks to a scheduler. During exploration, the algorithm explores and learns a representation of the behavior space through an AE trained online. 
    Any discovered reward is then exploited in the exploitation step through emitters.
    }
    \label{fig:stax}
\end{figure}

For an embodied agent whose goal is to learn a policy capable of solving a task, situations of \emph{sparse rewards} can be difficult to deal with.
The reason is that many policy-learning algorithms work by optimizing a \emph{reward function} providing feedback on the performances of the policy.
A well-designed reward function has to provide a reward often enough so the agent can know how good each performed action is \citep{sutton2018reinforcement}.
These kinds of rewards are called \emph{dense rewards}.
On the contrary, in sparse rewards settings, this feedback is provided sparingly, only after a given amount of time is passed or if a specific situation happens.
In these situations, it is difficult for a learning agent to evaluate how good a policy is and how appropriate each action is to each situation.
This can reduce the performance or even hinder the learning of a good policy.
%At the same time, sparse rewards settings are extremely common in situations in which the design of a proper reward function is costly or even impossible.
An example of this can be a robotic arm learning how to pick an object. 
The simplest way of rewarding the agent is to give the reward when the arm picks the object, while designing a reward that could lead the arm to pick the object is very hard.
For these reasons, when reward feedback is not readily available, a good strategy is to focus on \emph{exploration}, with the goal of finding a reward in the future.

\looseness=-1
Following this strategy, the way exploration is performed becomes fundamental.
Standard \gls{rl} algorithms, as described by \cite{sutton2018reinforcement}, perform exploration through random actions, a strategy that renders unlikely to find rewards if they are sparse enough.
This problem has been addressed with the introduction of different approaches, based on both \gls{rl} methods, \glspl{ea} or a mix of both \citep{sigaud2022combining}.
%\citep{Andrychowicz2017HER, Ecoffet2019GO_Explore, paolo2021sparse, hare2019dealing, riedmiller2018learning, Eysenbach2018DIAYN, lehman2008ns}.
Among them, \gls{ns} is an \gls{ea} that completely discards any performance information, focusing solely on exploration by looking for a set of policies whose behaviors are as different as possible \citep{lehman2008ns}.
This is done in a hand-designed space, the \gls{bs}, in which the behavior of each one of the generated policies is represented in order to evaluate their diversity.
The development of \gls{ns} has led to the birth of the evolution-based \emph{divergent search} family of algorithms, also known as \gls{qd} \citep{pugh2016qdfontier, cully2017quality}.
These methods, in addition to focusing on pure exploration through divergent search, can also optimize the performances of the discovered policies.
This grants a strong advantage over methods like \gls{ns} that tend to produce low-performing solutions with respect to the a posteriori evaluation on a rewarding task.
Nonetheless, the exploration abilities of these approaches, \gls{ns} included, are often limited by the need to hand-design \gls{bs}.
While this allows the designer to define what aspects of the problem need to be explored, it also increases the engineering cost of these methods while limiting the range of problems to which they can be applied.
To address this issue, researchers have introduced methods that can autonomously learn \gls{bs} through \emph{representation learning approaches}, thus reducing the amount of prior information needed for the design of the representation itself \citep{Liapis2013Delenox, paolo2020unsupervised, grillotti2021unsupervised}.
Supporting this approach, \cite{hagg2020analysis} have shown that autonomously learning \gls{bs} generates higher diversity of solutions compared to hand-designed \gls{bs}.
Notwithstanding the good results obtained by these methods, they are still limited either by the discarding of reward-related information of \gls{ns} \citep{Liapis2013Delenox, paolo2020unsupervised} or by the need to discretize the learned space \citep{grillotti2021unsupervised}.

\looseness=-1
\glsunset{name}
In this paper, we introduce the \gls{name} algorithm, a method that can perform exploration in a search space that is autonomously learned at execution time, while also optimizing any possible discovered reward. 
As with \gls{ns}, this exploration is completely reward-agnostic, but contrary to this method, once an area of the search space is discovered to contain a reward, \gls{name} performs a local search in this area with the goal to optimize the total obtained reward.
This optimization is performed through \emph{emitters}, a concept introduced by \cite{fontaine2020covariance}, consisting of instances of reward-based \glspl{ea} used to perform local search in an area of the whole \gls{bs}.
The idea of emitters was used in \gls{serene} \citep{paolo2021sparse} to optimize any reward discovered during the search performed by \gls{ns}.
At the same time, \gls{serene} still relies on a hand-designed behavior space in which to perform the search. 
\gls{name} builds on \gls{serene} by removing this requirement through the use of an \gls{ae} to learn the behavior space online while performing the search \cite{grillotti2021unsupervised, paolo2020unsupervised}.

\looseness=-1
\glsreset{name}
\gls{name} deals with sparse reward problems by separating the exploration and the learning of the unknown search space from the exploitation of any possible reward through an alternating two-step process.
In the first step, the algorithm explores the search space guided by the low-dimensional representation of the policies behavior given by the \gls{ae}.
At the same time, this representation is learned by training the \gls{ae} on the data collected during the evaluation of the discovered policies.
When rewards are found, they are exploited in the second step through the use of \emph{emitters}, in a way similar to \gls{serene} \citep{paolo2021sparse}.
The clear separation between exploration and exploitation has many advantages.
The two processes often push the optimization in different directions: exploration requires trying as many things as possible, while exploitation requires getting better at the things we already know.
Working on them separately, then allows doing both without degrading performances, as it can happen in multi-objective approaches like NSGA-II \citep{paolo2021sparse}.
Moreover, the decoupling of exploration from exploitation enables using different strategies for either of the two processes in a more modular approach.

\looseness=-1
To recap, \gls{name} performs three main tasks: (1) learning a behavior space while (2) exploring it, and (3) efficiently exploiting a reward once it is found.
The method builds on \gls{ns} by adding an \gls{ae} to learn a low dimensional representation of the search space.
Moreover, the reward is exploited through emitters to quickly improve on rewards.
The advantages provided by \gls{name} are twofold: 
(1) it can deal with sparse rewards situations by discovering and quickly optimizing the rewards, thanks to the separation of the exploration process from the exploitation of the reward provided by the use of emitters;
(2) by autonomously learning the \glsfirst{bs}, it removes the limitation of classical divergent-search approaches requiring a hand-designed space, thus reducing the amount of prior information needed at design time.
All of this allows \gls{name} to deal with sparse reward environments with minimum prior information required about the task at design time.

\looseness=-1
The paper is organized as follows:
Sec. \ref{sec:background} will present an overview of related work and the methods on which \gls{name} builds.
The \gls{name} method itself is detailed in Sec. \ref{sec:method}, while the experimental settings on which it has been tested are shown in Sec. \ref{sec:experiments}.
The obtained results are shown and discussed in Sec. \ref{sec:stax_results}.
The paper concludes with Sec. \ref{sec:stax_conclusion}, in which possible extensions and improvements are discussed.
\looseness=-1

\section{Background and related work}
\label{sec:background}
\looseness=-1
This section presents an overview of other works on the sparse rewards problem, together with an explanation of how \gls{ns} and \emph{emitters} work.

\subsection{Sparse Reward}
\looseness=-1
For many policy learning approaches, the reward function is fundamental: it is through this function that the designer communicates to the learning agent what is the goal the policy should solve \citep{sutton2018reinforcement}.
If the reward signal is given sparingly, after a lot of time, or only if certain conditions are met, the agent can often find itself in situations in which no reward is present, thus with no signal to drive the learning.
To address this issue, many approaches have been proposed.
Some of these approaches rely on \emph{reward shaping} \citep{mataric1994reward, ng1999policy}, a technique consisting of augmenting the original reward function with additional features that are supposed to provide the agent with better guidance in solving the task \citep{hu2020learning, berner2019dota, trott2019keeping}.
Another successful strategy is the self-assigning of goals.
This can be done either by using information from previously encountered situations \citep{Andrychowicz2017HER}, or by using the representations of an unsupervised learning algorithm over a distribution of possible targets \citep{Nair2018ImaginedGoal}.

\looseness=-1
A different approach is based on \gls{im} \citep{oudeyer2009intrinsic, aubret2019survey}, by having the agent generate its own learning signal, without the need for any environmental reward.
This can be obtained by estimating the novelty of a state by considering how often that state has been visited \citep{bellemare2016unifying, burda2018exploration}.
The less novel a state is, the more the agent is pushed to go elsewhere, thus performing more exploration.
\gls{gep} are another family of algorithms that use the self-assignment of goals to foster exploration \citep{baranes2013active, Forestier2017IMGEP, laversanne2018curiosity}.
\cite{Forestier2017IMGEP} use this to first learn a goal-parametrized policy and then use this policy to solve the given task.
These approaches have also been used with two-phase strategies to help separate the exploration process from the exploitation of the possible discovered rewards \cite{colas2018gep, Ecoffet2019GO_Explore}.

\looseness=-1
\emph{Divergent-search algorithms} are a family of \glspl{ea} specifically designed to focus on exploration, rendering them naturally suited to deal with sparse reward situations \citep{lehman2008ns, cully2017quality, pugh2016qdfontier}.
The first introduced method of this family is \gls{ns}, introduced by \cite{lehman2008ns}, which works by completely ignoring any reward signal in order to generate a set of solutions as diverse as possible.
Inspired by \gls{ns}, many other methods have been introduced that not only focus on the diversity of the solutions but also optimize their performances with respect to a given objective.
This gave rise to a new family of methods called \glsfirst{qd} \citep{cully2017quality, pugh2016qdfontier, CUlly2015MAPElites, Eysenbach2018DIAYN, lehman2011evolving, paolo2021sparse, mouret2015illuminating}.
Moreover, given the great exploration abilities provided by divergent-search algorithms, some researchers combined them with \gls{rl} methods to better deal with sparse reward situations \citep{colas2018gep, cideron2020qd}.

\subsection{Novelty Search}
\label{sec:ns}
\looseness=-1
\glsfirst{ns} is an \gls{ea} that drives the search by focusing on maximizing the diversity of a set of solutions \citep{lehman2008ns}.
To do this, the algorithm uses a metric called \emph{novelty}, calculated in an \emph{hand-designed behavior space} $\mathcal{B}$ in which the behavior of each policy is represented.
This space, in the literature also called \emph{outcome space} \citep{paolo2020billiard}, is at the heart of \gls{ns} and needs to be tailored to the problem at hand by using prior knowledge of the system and the task.

\looseness=-1
The algorithm works by evaluating each policy, parametrized by a set of parameters $\theta_i \in \Theta$, on the system for $T$ time-steps.
During this evaluation, the system traverses a set of states $s_t$ generating a trajectory of traversed states $\tau_s = [s_0, \dots, s_T]$.
These states are observed by the agent through some sensors, generating a corresponding trajectory of observations $\tau_\mathcal{O} = [o_0, \dots, o_T]$, where $o_t \in \mathcal{O}$ is the, possibly partial, observation of state $s_t$.
These observations can be generated in different ways, depending on the setting.
If the states are known, the agent can directly work with them, in which case $o_t = s_t$.
In other situations, the state needs to be observed through sensors, in which case $o_t$ would be a, possibly partial, representation of $s_t$.
The trajectory of observations $\tau_\mathcal{O}$ is then used to extract the corresponding behavior descriptor $b_i \in \mathcal{B}$ of the policy $\theta_i$ through an observer function $O_\mathcal{B} : \mathcal{O} \rightarrow \mathcal{B}$.
The whole process is summarized by using a \emph{behavior function} $\phi : \Theta \rightarrow \mathcal{B}$ that directly maps a policy to its behavior descriptor:
\begin{equation}
    \phi(\theta_i) = b_i.
    \label{eq:behav_func}
\end{equation}
Once the behavior descriptors of all the policies in a population have been calculated, the novelty of a policy $\theta_i$ in the population can be obtained by measuring the average distance of its behavior descriptor with respect to the descriptors of its $k$ closest policies.
The higher this distance, the more novel the behavior of a policy is considered.
The novelty $\eta(\theta_i)$ is calculated through the following equation:
\begin{equation}
    \eta(\theta_i) = \frac{1}{|J|} \sum_{j \in J} \text{dist}(b_i, b_j) = \frac{1}{|J|} \sum_{j \in J} \text{dist}(\phi(\theta_i), \phi(\theta_j))
    \label{eq:novelty}
\end{equation}
where $J$ is the set of indexes of the $k$ closest policies to $\theta_i$ in $\mathcal{B}$.

\looseness=-1
At each generation, the novelty of the policies is calculated and the ones with the highest novelty are selected to be part of the next generation population.
At the same time $N_Q$ policies are selected at each generation to be stored into an \emph{archive} $\mathcal{A}_{\text{Nov}}$.
The function of the archive is to keep track of the already explored areas of the search space, pushing the search towards less visited areas.
This is done by selecting the $|J|$ policies used for the novelty calculation in equation \ref{eq:novelty} not only from the current population but also from the archive.

\looseness=-1
% \gls{ns} tends to uniformly cover the search space \citep{doncieux2019ns_theory}, property that renders it a good candidate for exploration in sparse rewards settings.
% Notwithstanding this, \gls{ns} has two main limitations, the biggest of which is the requirement for the outcome space $\mathcal{B}$ to be hand-designed. 
% This is not always feasible and in many situations it requires a huge engineering effort.
% The second limitation is that \gls{ns} completely ignores any information related to the reward.
% This means that the algorithm can only perform exploration, but not optimize the performances of any solution.
% In situations of sparse rewards this can prove a big limitation, because even if sparse, rewards can render the search more efficient.

\subsection{Learning a behavior descriptor}
\looseness=-1
At the core of many divergent search algorithms lies a hand-designed \glsfirst{bs}.
The need to hand-design this space poses strong limitations for the application of these methods to various problems in which the factors important for the exploration are not clear.
To overcome this problem, many approaches that use representation learning methods to learn a low-dimensional representation of the behavior of the policy have been recently proposed \citep{paolo2020unsupervised, cully2019autonomous, Liapis2013Delenox}.

\looseness=-1
\cite{cully2019autonomous} uses the learned low-dimensional representation to describe the behavior of the policy and select in which cell of the MAP-Elites grid the policy itself belongs.
At the same time, \gls{taxons} \citep{paolo2020unsupervised} selects the policies not only based on the novelty calculated through the learned low-dimensional representation but also on the reconstruction error of the \gls{ae} through a metric called \emph{surprise} \citep{gaier2019quality}.
The idea behind this is that the higher the reconstruction error, the less often a behavior has been seen, thus the more novel it is.
This is similar to the approaches introduced by \cite{burda2018exploration} and \cite{salehi2021br}.
\gls{name} uses \gls{taxons} to learn the low-dimensional representation of the behavior of a policy during the exploration phase, thus removing the need to hand-design \gls{bs}.
At the same time, rather than selecting the policies according to only one of the two metrics, novelty or surprise, as done by \gls{taxons}, it uses the NSGA-II \gls{moo} approach \citep{deb2002fast} to combine both objectives and select the policies at each generation.

\subsection{Emitters}
% \looseness=-1
% Notwithstanding its exploration capabilities, vanilla \gls{ns} is not equipped to take advantage of any reward that can be found during the search. 
% This limits the power of the algorithm and discards some important information on the task that can be used to steer and improve the efficiency of the search.
% Many solutions have been proposed to address this problem (\cite{lehman2011evolving, mouret2015illuminating, CUlly2015MAPElites, cully2020multi, paolo2021sparse}), leading to the development of the \gls{qd} family of algorithms (\cite{pugh2016qdfontier, cully2017quality}).

\looseness=-1
Among \gls{qd} algorithms worth of notice are approaches using \emph{emitters}.
Introduced by \cite{fontaine2020covariance} and later used by \cite{cully2020multi} and \cite{paolo2021sparse}, emitters are instances of local-search reward-based \glspl{ea} instantiated during the global search performed by another \gls{ea}, allowing the quick exploration of a small area of the search space while optimizing on the reward.
There is no limitation on the kind of algorithm to use as an emitter.
In the work from \cite{fontaine2020covariance}, the CMA-ME algorithm uses MAP-Elites in conjunction with estimation-of-distribution emitters.
The algorithm works by sampling a policy $\theta_i$ from the MAP-Elites archive and using it to initialize the population of an emitter $\mathcal{E}_i$, which is then evaluated until a termination condition is met.
The policies discovered are added to the MAP-Elites archive according to a given addition strategy. 
Once an emitter is terminated, another policy $\theta_j$ is selected from the MAP-Elites archive to initialize another emitter.
The cycle is repeated until the whole evaluation budget is depleted.
% The main limitation in using estimation-of-distribution \gls{ea} as emitter instances is due to the size of the emitter population compared to the dimension of parameter space $\Theta$. 
% If the population is smaller than $|\Theta|$, the estimation of the covariance matrix of the distribution $\Sigma$ can be unreliable.
% CMA-ES overcomes this problem by taking into account information from previous generations when calculating $\Sigma$, but this can lead to a less efficient use of the exploration budget.

\looseness=-1
Another method using emitters is \gls{serene}.
Introduced by \cite{paolo2021sparse}, the algorithm is based on \gls{ns} and targets explicitly sparse rewards problems.
Contrary to CMA-ME, \gls{serene} works through an alternating two-stage process, one performing exploration, the other exploiting the found rewards.
Exploration is done through \gls{ns} over the hand-designed \glsfirst{bs}.
Once a reward is discovered, it is exploited in the exploitation step when emitters are launched over the rewarding area of the search space $\mathcal{B}_R \subseteq \mathcal{B}$.
This two-steps process allows the algorithm to easily deal with sparse rewards settings in which, while the search can be global, the optimization of the reward has to be local around the rewarding policy.

\looseness=-1
The method introduced in this work augments \gls{serene} with the ability to autonomously learn \gls{bs} through a strategy similar to \gls{taxons} \citep{paolo2020unsupervised}.
In the next sections we will detail how \gls{name} works and how, by taking advantage of emitters and the unsupervised learning of the behavior space, it is possible to quickly explore an unknown search space while efficiently optimizing any possible discovered reward.
\section{Method}
\label{sec:method}
\looseness=-1
\gls{name} deals with \emph{sparse rewards} settings by separating the search process into two alternating sub-processes: one performing \emph{exploration} of the search space and another performing \emph{exploitation} of any discovered reward.
The algorithm starts with the \emph{exploration} phase and then the two processes are alternated through a \emph{meta-scheduler}.
The task of the meta-scheduler is to split the total evaluation budget $Bud$ in small chunks of size $K_{Bud}$ and assign them to either one of the two sub-processes.
In the \emph{exploration} phase, \gls{name} learns a \glsfirst{bs} from high dimensional observations of the environment through an \gls{ae}, and explores it.
Meanwhile, during the \emph{exploitation} phase, the discovered rewarding policies are optimized  through \emph{emitters}, instances of local-search reward-based optimization algorithms. 
Note that, while in this work we use an \emph{elitist \gls{ea}}, any kind of optimization algorithm can be used as emitter.
Once the whole evaluation budget $Bud$ is depleted, the algorithm returns two collection of policies: the \emph{novelty archive} $\mathcal{A}_\text{Nov}$, containing the diverse-behavior policies found during the exploration phase, and the \emph{reward archive} $\mathcal{A}_\text{Rew}$, containing the reward-optimized policies found during the exploitation phase.
The whole process is designed in a way that allows the discovery of different high-reward policies with minimal prior information about the task.

\looseness=-1
There are two aspects of \gls{name} that are worth highlighting: the autonomous learning and exploration of the behavior space and the optimization of the reward through emitters.
Autonomously learning the \glsfirst{bs} allows to reduce the amount of prior information needed to solve the task by removing the need to hand-design \gls{bs}.
This is achieved by learning a low-dimensional representation of this space through an \gls{ae}, directly from high-dimensional observations collected during the policy evaluation, in a fashion similar to \gls{taxons} \citep{paolo2020unsupervised}.
The search is then driven by using the information extracted by the \gls{ae} from the observations collected during the evaluations of the policies.
The encoder part of the \gls{ae} can in fact be used as \emph{observation function} and its \emph{latent feature space} $\mathcal{F}$ as \glsfirst{bs}.
The behavior descriptor of a policy is obtained by sampling multiple high-dimensional observations along its trajectory and use the \gls{ae} to extract a compressed representation.
% At the same time, while in \gls{taxons} the authors made the assumption that the final frame of the trajectory was informative enough to extract the behavior of the policy from it, in \gls{name} we remove this assumption.
% To do this, \gls{name} samples multiple frames along the generated trajectory and uses the corresponding \gls{ae} representations to generate the behavior descriptor.
Moreover, as for the first iterations of the search the \gls{ae} representation does not properly represent the behavior space yet \citep{grillotti2021unsupervised}, the training happens more frequently.
Once a few training iterations have been performed, the \gls{ae} can better represent the behaviors, so the training happens less and less frequently.
A detailed description of the training process of the \gls{ae} is given in Sec.~\ref{sec:ae_training}.

\looseness=-1
The second important aspect of \gls{name} is the optimization of the reward through emitters. 
If, during the exploration phase, a policy $\theta_i$ obtains a reward, it will be used during the exploitation phase to instantiate an emitter in order to improve on the reward.
The rationale is that behaviors similar to the rewarding behavior $\psi(\theta_i)$ are likely rewarding too, with possibly even higher performances than $\psi(\theta_i)$.
These behaviors can be considered belonging to the subspace of rewarding behaviors $\mathcal{B}_{Rew} \in \mathcal{B}$ and their corresponding policies can be discovered by performing local search around $\theta_i$ through emitters.
Note that the reward exploitation performed during this phase does not rely on any behavior descriptor.
The quality of \gls{bs} learned representation then does not interfere with the reward optimization process.
This means that if a reward is discovered at the initial stages of the search, when behavior space has not been learned yet, \gls{name} can still exploit it thanks to descriptor-less emitters, without loss of performance.
The details of the reward optimization are given in Sec.~\ref{sec:exploitation}.
\looseness=-1
\setlength{\textfloatsep}{0.2cm}
\begin{algorithm}[!t]
\caption{\gls{name}}\label{alg:main}
\textbf{INPUT:} evaluation budget $Bud$, budget chunk size $K_{Bud}$, population size $M$, emitter population size $M_{\mathcal{E}}$, offspring per policy $m$, mutation parameter $\sigma$, number of policies added to novelty archive $Q$, \gls{ae} training interval $TI$, randomly initialized \gls{ae}, number of bootstrap generations $\lambda$\;
\textbf{RESULT:} Novelty archive $\mathcal{A}_{\text{Nov}}$, rewarding archive $\mathcal{A}_{\text{Rew}}$, trained \gls{ae}\;
$\mathcal{A}_{\text{Nov}} = \emptyset$,
$\mathcal{A}_{\text{Rew}} = \emptyset$,
$\mathcal{Q}_{\text{Em}} = \emptyset$,
$\mathcal{Q}_{\text{Cand\_Nov}} = \emptyset$,
$\mathcal{Q}_{\text{Cand\_Em}} = \emptyset$,
$D = 0$,
$TI_C = 0$\;
Sample initial population $\Gamma_0$\;
Split $Bud$ in chunks of size $K_{Bud}$\;
\While{$Bud$ not depleted}{
    \If{$\Gamma_0$}{
        \emph{Eval}($\theta_i), ~~ \forall \theta_i \in \Gamma_0$; \CommentSty{\SLASH\SLASH Evaluate initial population}
    $b_i = \psi(\theta_i) \in \mathcal{B}, ~~ \forall\theta_i \in \Gamma_0$; ~~ \CommentSty{\SLASH\SLASH Calculate behavior descriptor}\
    }
    \emph{Exploration}($K_{Bud}$, $m$, $\sigma$, $\mathcal{A}_{\text{Nov}}$, $\mathcal{Q}_{\text{Cand\_Em}}$, $\Gamma_g$, $Q$, \gls{ae}); ~~ \CommentSty{\SLASH\SLASH Alg.~\ref{alg:stax_exploration}}\\
    $TI_C = TI_C + 1$; ~~ \CommentSty{\SLASH\SLASH Increase training interval counter} \\
    \CommentSty{/* Update autoencoder and descriptors */}\\
    \If{$TI_C == TI$}{
    $DS = $ \emph{Extract\_Dataset}($\mathcal{A}_{\text{Nov}}$, $\mathcal{A}_{\text{Rew}}$, $\Gamma_g$, $\Gamma^m_g$)\;
    \emph{Train\_Autoencoder}(\gls{ae}, $DS$)\;
    \emph{Update\_Descriptors}(\gls{ae}, $\Gamma_g$, $\Gamma^m_g$, $\mathcal{A}_{\text{Nov}}$, $\mathcal{A}_{\text{Rew}}$, $\mathcal{Q}_{\text{Em}}$, $\mathcal{Q}_{\text{Cand\_Nov}}$, $\mathcal{Q}_{\text{Cand\_Em}}$)\;
    $TI = TI + 1$\;
    $TI_C = 0$\;
    }

    \CommentSty{/*  If rewarding policies have been found */}\\
    \If{$\mathcal{Q}_{\text{Cand\_Em}} != \emptyset$ \textbf{or} $\mathcal{Q}_{\text{Em}} != \emptyset$}{
    \emph{Exploitation}($K_{Bud}$, $\mathcal{Q}_{\text{Cand\_Em}}$, $\lambda$, $m$, $\mathcal{Q}_{\text{Em}}$, $\mathcal{A}_{\text{Nov}}$, $\mathcal{A}_{\text{Rew}}$, $M_{\mathcal{E}}$); ~~ \CommentSty{\SLASH\SLASH Alg.~\ref{alg:exploitation}}\\
    }
}
\end{algorithm}

\begin{figure}[!h]
    \centering
    \includegraphics[width=.8\linewidth]{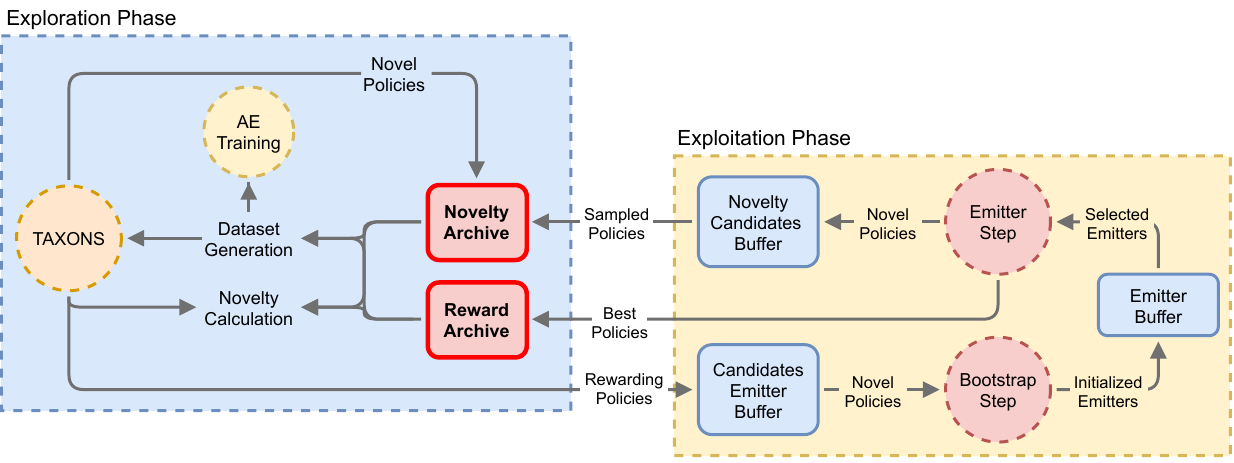}
    \vspace*{-2mm}
    \caption{Overview of the containers used during the search by \gls{name} to track the discovered policies and the initialized emitters. The two collections returned as outputs of the algorithms are highlighted in red.
    }
    \label{fig:sets}
\end{figure}

\looseness=-1
During its operation, \gls{name} tracks the policies generated in the different phases of the search through the following buffers and containers:
\begin{itemize}
    \item \emph{novelty archive} $\mathcal{A}_\text{Nov}$: a collection of policies with diverse behaviors found during the \emph{exploration phase}. One of the two collections of policies returned as outputs of \gls{name};
    \item \emph{reward archive} $\mathcal{A}_\text{Rew}$: a collection of the most rewarding policies found during the \emph{exploitation phase}. Other collection of policies returned as outputs of \gls{name};
    \item \emph{candidates emitter buffer} $\mathcal{Q}_\text{Cand\_Em}$: a buffer in which the rewarding policies $\psi(\theta_i) \in \mathcal{B}_\text{Rew}$ found during the \emph{exploration phase} are stored before being used to initialize emitters in the \emph{exploitation phase};
    \item \emph{emitter buffer} $\mathcal{Q}_\text{Em}$: a buffer in which the initialized emitters to be evaluated during the \emph{exploitation phase} are stored;
    \item \emph{novelty candidates buffer} $\mathcal{Q}_\text{Cand\_Nov}$: an emitter-specific buffer in which the most novel policies found by the emitter are stored.
    Each emitter has its own novelty candidate buffer from which policies are sampled to be added to $\mathcal{A}_\text{Nov}$ at the termination of the emitter itself.
\end{itemize}
A high-level overview of the interactions between the buffers and containers is shown in Fig. \ref{fig:sets}.
\setlength{\textfloatsep}{0.2cm}

\looseness=-1
The three main steps of \gls{name} - exploration, training of the \gls{ae} and exploitation of the reward - are detailed respectively in sections \ref{sec:exploration}, \ref{sec:ae_training} and \ref{sec:exploitation}.
The whole \gls{name} algorithm is illustrated in Fig. \ref{fig:stax} and described in Alg. \ref{alg:main}.

\subsection{Exploration} 
\label{sec:exploration}
\looseness=-1
Having minimal prior information about the task, \gls{name} starts by performing the exploration step.
The first time this step is performed, the parameters $\theta \in \Theta$ of the $M$ policies in the initial population $\Gamma_0$ randomly sampled from the normal distribution $\mathcal{N}(0, \mathbb{I})$, as the policies are \gls{nn}.
Note that any parametric function $f(s|\theta) = a$, where $s$ is the state of the system and $a$ is the action, can be used as policy.
The weights of the \gls{ae} used to drive the search are also randomly sampled.
At each generation $g$, $m$ policies $\theta_i^j$ are generated for each policy $\theta_i$ in the current population $\Gamma_g$ through a \emph{mutation operator}.
This will result in an offspring population $\Gamma^m_g$ of size $m \times M$ whose policies are formed as:
\begin{equation}
    \forall j \in \{1, \dots, m\}, \forall i \in \{1, \dots, M\}, \theta_i^j = \theta_i + \epsilon, ~\text{with}~ \epsilon \sim \mathcal{N}(0, \sigma I).
    \label{eq:offsprings}
\end{equation}
The policies in the offspring population $\theta \in \Gamma^m_g$ are then evaluated.
During the evaluation of a policy $\theta_i$ the system traverses a trajectory of states $\tau^i_s = [s^i_0, \dots, s^i_T]$ that are observed through sensors, generating a corresponding trajectory of observations $\tau^i_o = [o^i_0, \dots, o^i_T]$. 
The policy is then assigned a behavior descriptor $b_i$ obtained by using \emph{multiple observations} sampled along $\tau^i_{\mathcal{O}}$.
The descriptor is generated by encoding the sampled observations thanks to the \gls{ae}'s encoder $E(\cdot)$ and then stacking their low-dimensional representations together.
\begin{algorithm}[!t]
\caption{\gls{name} Exploration Phase}\label{alg:stax_exploration}
\textbf{INPUT:} budget chunk $K_{Bud}$, number of offspring per parent $m$, mutation parameter $\sigma$, novelty archive $\mathcal{A}_{\text{Nov}}$, candidate emitters buffer $\mathcal{Q}_{\text{Cand\_Em}}$, population $\Gamma_g$, number of policies $N_Q$, autoencoder \gls{ae}\;
\While{$K_{Bud}$ not depleted}{
    Generate offspring $\Gamma^m_g$ from population $\Gamma_g$\;
    \CommentSty{/* Loop over the policies in the population */}\\ 
    \For{$\theta_i \in \Gamma^m_g$}{
        \emph{Eval}($\theta_i$); ~\CommentSty{\SLASH\SLASH Evaluate policy}\\
        $b_i=\psi(\theta_i)=[\dots, E(o^i_{t_k}), \dots, E(o^i_{t_K})]$;~ \CommentSty{\SLASH\SLASH Calc. behavior descr.}\\
    }
    \For{$\theta_i \in \Gamma^m_g$}{
        $\eta(\theta_i) = \frac{1}{|J|}\sum_{j \in J}\text{dist}(b_i, b_j)$; ~~ \CommentSty{\SLASH\SLASH Calculate novelty}\\
        $s(\theta_i) = \sum_{k \in K}\big|\big|o_{t_k}^{i} - D\big(E(o_{t_k}^{i})\big)\big|\big|^2$;  ~~ \CommentSty{\SLASH\SLASH Calculate surprise}\\
        \CommentSty{/* If the policy has a rewarding behavior */}\\
        \If{$\psi(\theta_i) \in \mathcal{B}_{\text{Rew}}$}{
            $\mathcal{Q}_{\text{Cand\_Em}} \leftarrow \theta_i$;  ~~ \CommentSty{\SLASH\SLASH Store rewarding policy}\\
        }
    }
    $\mathcal{A}_{\text{Nov}} \leftarrow Sample(\Gamma^m_g, N_Q)$;  ~~ \CommentSty{\SLASH\SLASH Store most novel $N_Q$ policies}\\ 
    \CommentSty{/* NSGA-II policy selection wrt novelty and surprise */}\\
    Calculate non dominated fronts $F_j, ~~ \forall\theta_i \in \Gamma^m_g \bigcup \Gamma_g$\;
    Sort fronts according to \emph{non domination}\;
    Generate $\Gamma_{g+1}$ from most non dominated solutions $\theta_i \in F_j$\;
    \If{If last front $F_J$ is partially selected}{
        Calculate \emph{crowding distance} $\forall\theta_i \in F_J$\;
        Complete filling up $\Gamma_{g+1}$ with less crowded solution $\theta_i \in F_J$\;
    }
}
\end{algorithm}
This can be described as $b_i = \psi(\theta_i) = [\dots, E(o^i_{t_k}), \dots, E(o^i_{t_K})]$,
where $o^i_{t_k}$ is the observation generated by the policy $\theta_i$ at time-step $t_k$.
Sampling multiple observations along the trajectory is in contrast to what \cite{paolo2020unsupervised} did in \gls{taxons}, in which only the last observation was used to generate the descriptor.
Using the last observation in fact requires such observation to be informative of the behavior of the policy over the whole trajectory.
On the contrary, using multiple observations along $\tau_s$, such an assumption is not required anymore.

\looseness=-1
The diversity of a policy is evaluated through two metrics: \emph{novelty} and \emph{surprise}.
The first one is the normalized euclidean distance in the learned \gls{bs}:
\begin{equation}
    \eta(\theta_i) = \frac{1}{|J|} \sum_{j \in J} \text{dist}(\psi(\theta_i), \psi(\theta_j)).
    \label{eq:ae_novelty}
\end{equation}
At the same time, the surprise is calculated as the \emph{sum} of the \gls{ae}'s \emph{reconstruction error} over each one of the sampled observations generated by $\theta_i$.
A higher surprise implies that the \gls{ae} has not seen that area of the learned behavior space very often.
This means that selecting policies with high surprise leads the algorithm to increased exploration.
Such metric is defined as:
\begin{equation}
\label{eq:stax_surprise_ae}
    s(\theta_i) = \sum_{k \in K}\big|\big|o_{t_k}^{i} - D\big(E(o_{t_k}^{i})\big)\big|\big|^2,
\end{equation}
where $K$ is the list of indexes of the selected time-steps along the trajectory.

\looseness=-1
The two metrics are used to select the policies that will form the population for the next generation $\Gamma_{g+1}$ through the NSGA-II multi-objective approach \citep{deb2002fast}.
This is in contrast to what was done by \cite{paolo2020unsupervised} in \gls{taxons}, in which only one among novelty and surprise was sampled at each generation to be used for policy selection.

\looseness=-1
Finally, \gls{name} samples uniformly $N_Q$ policies to be added to the \emph{novelty archive} $\mathcal{A}_{\text{Nov}}$.
Moreover, all the rewarding policies found in this phase are added in the \emph{candidates emitter buffer} $\mathcal{Q}_{\text{Cand\_Em}}$ to be used during the \emph{exploitation phase} to generate emitters.
The whole exploration process is shown in Algorithm \ref{alg:stax_exploration}.

\subsection{Training of the autoencoder}
\label{sec:ae_training}
\looseness=-1
The exploration performed by \gls{name} is driven by the \gls{ae}.
This means that the way the \gls{ae} itself is trained, and thus the quality of the learned low-dimensional representation, is fundamental in order to obtain good exploration.  
In order to meaningfully look for diversity in the learned behavior space $\mathcal{B}$, the \gls{ae} has to be trained on the data collected during the search for policies itself.
This data is collected into a dataset $DS$ consisting of the observations used to generate the behavior descriptor of the policies, as defined in Sec.~\ref{sec:exploration}.
The policies whose observations are added to $DS$ are the ones contained in both the \emph{reward archive} $\mathcal{A}_{\text{Rew}}$ and the \emph{novelty archive} $\mathcal{A}_{\text{Nov}}$, with the addition of the observations from the population $\Gamma_g$ and the offspring population $\Gamma_g^m$ of the last evaluated generation $g$.
The data of the archives provides a \emph{curriculum}, stabilizing the training process and preventing the search from cycling back to already explored areas.
% Additionally, the data from the $\mathcal{A}_{\text{Rew}}$ policies helps stabilizing the training process. 
% These policies are in fact selected not according to the novelty calculated thanks to the \gls{ae}, but only for their reward, that is a factor independent from the feature space.
At the same time, adding the observations from the most recent population to the training dataset helps the \gls{ae} to better represent the frontier of the explored space, towards which the search is to be pushed.

\looseness=-1
Once the dataset $DS$ has been collected, it is split into two sub-datasets: the \emph{training dataset} $DS_{\text{Train}}$ and the \emph{validation dataset} $DS_{\text{Val}}$.
For each training episode, the \gls{ae} is trained on the $DS_{\text{Train}}$.
At the end of each training epoch on $DS_{\text{Train}}$, the model validation error is calculated on $D_{\text{Val}}$.
The training episode is stopped if the error increases for 3 consecutive epochs.

\looseness=-1
As stated in Sec. \ref{sec:method}, the \gls{ae} is trained less frequently the longer the search is performed; the same strategy is employed in the AURORA method \citep{cully2019autonomous}.
This allows adapting the frequency of the training to the maturity of the learned \gls{bs}, while saving time and computational resources with respect to training the \gls{ae} at fixed intervals. 
%In fact, after the first training episodes, the learned representation is mature enough for the \gls{ae} to start focusing on its refinement.
%This allows to train the \gls{ae} less often, allowing to save time and computational resources.
% Moreover, by training less frequently, the possible overfitting of the \gls{ae} on the data present in the archives is limited.
This shifting training regime is obtained by performing the training process every $TI$ exploration steps.
At the beginning of the search, \gls{name} sets $TI = 1$.
Its value is then increased by $1$ every time a training episode is performed.
Finally, at the end of each training episode, the behavior descriptor of all the policies present in the archives and in the populations is updated with the new descriptors generated by the retrained \gls{ae}.
This keeps the behavior descriptors and the novelty measurements of the policies consistent and meaningful.

\subsection{Exploitation}
\label{sec:exploitation}
\looseness=-1
At the end of the exploration step, if the \emph{emitters candidate buffer} $\mathcal{Q}_\text{Cand\_Em}$ or the \emph{emitters buffer} $\mathcal{Q}_\text{Em}$ are not empty, the meta-scheduler assigns a budget chunk $K_{Bud}$ to the exploitation step.
The objective of this phase is to optimize the reward.
In practice, this consists of i) identifying the policies that can be used to initialize the populations of the emitters and ii) running such emitters with the goal of generating solutions with high rewards.
This is done through two sub-steps: the \emph{bootstrap step} and the \emph{emitter step}.
During the bootstrap step, the rewarding policies collected in the \emph{emitters candidate buffer} $\mathcal{Q}_\text{Cand\_Em}$ are used to instantiate emitters that are then evaluated for few iterations.
The emitter with the potential to improve on the reward are added to the \emph{emitter buffer} $\mathcal{Q}_\text{Em}$ to be fully evaluated during the subsequent emitter step.
At the same time, the emitters not capable of improving on the reward are discarded.
In the following sub-step, the emitters from $\mathcal{Q}_\text{Em}$ are sampled according to their performance and evaluated until termination or until the budget chunk is depleted.

\looseness=-1
Such a two sub-steps process allows \gls{name} to quickly decide which of the rewarding policies from $\mathcal{Q}_\text{Cand\_Em}$ is worth optimizing and which is not. 
This prevents the waste of computational budget on the optimization of policies that are in hard-to-escape local optima of the reward landscape.
Moreover, using emitters allows to disjointly optimize multiple reward areas in an efficient way by quickly finding good solutions.
This is fundamental for an approach like \gls{name} in which the \glsfirst{bs} is autonomously learned.
In hand-designed \gls{bs} the engineer has total control over the space itself, allowing him to reduce the disjointedness of the reward areas.
This is not the case when the behavior descriptor is generated by stacking multiple learned representations extracted from high-dimensional observations, as done by \gls{name}.
In this kind of setting, there is no guarantee that the new space will have the same structure of the reward areas as the ground-truth hand-designed \gls{bs}. 
Given the complex nature of the learned space, due to the stacking of the encoding of multiple observations, it can happen that this space contains multiple reward areas, even if only one is present in the ground-truth \gls{bs}.
For these reasons, using an emitter-based approach as \gls{name} capable of focusing on multiple reward areas can give a strong advantage in situations where the behavior representation is so complex.

\looseness=-1
In the following, we will describe in detail first how our emitters work and how the two sub-steps of the exploitation process use them to optimize the reward.
The whole exploitation phase is detailed in Algorithm \ref{alg:exploitation}.

\begin{algorithm}[!t]
\caption{\gls{name} Exploitation Phase}\label{alg:exploitation}
\textbf{INPUT:} budget chunk $K_{Bud}$, candidate emitters buffer $\mathcal{Q}_{\text{Cand\_Em}}$, number of bootstrap generations $\lambda$, emitter population size $M_{\mathcal{E}}$, number of offspring per policy $m$, emitters buffer $\mathcal{Q}_{\text{Em}}$, rewarding archive $\mathcal{A}_{\text{Rew}}$, novelty archive $\mathcal{A}_{\text{Nov}}$\;
\CommentSty{/* Bootstrap step */}\\
\While{$\nicefrac{K_{Bud}}{3}$ not depleted}{
Select most novel policy $\theta_i$ from $\mathcal{Q}_{\text{Cand\_Em}}$\;
\CommentSty{\SLASH\SLASH Calculate emitter's mutation standard deviation}\\
$\sigma_i = \nicefrac{\min_j(\text{dist}(\theta_i, \theta_j))}{3}, ~~ \forall\Tilde\theta_j \in \Gamma_g^m \cup \Tilde\Gamma_g$\; 
Initialize: $\mathcal{E}_i$, $\mathcal{Q}^i_{\text{Cand\_Nov}} = \emptyset$, and $P_0$\;
\For {$\gamma \in \{0, \dots, \lambda\}$}{
    \If{$P_0$}{
        \emph{Eval}($\Tilde{\theta}_j$), $\forall \Tilde{\theta}_j \in P_0$; \CommentSty{\SLASH\SLASH Evaluate initial population}\\
    }
    Generate offspring population $P^m_{\gamma}$ from $P_{\gamma}$\;
    \emph{Eval}($\Tilde{\theta}_j$), $\forall \Tilde{\theta}_j \in P^m_{\gamma}$; \CommentSty{\SLASH\SLASH Evaluate offspring population}\\
    Generate $P_{\gamma+1}$ from best $\Tilde{\theta}_j \in P^m_{\gamma} \bigcup P_{\gamma}$\;
}
Calculate improvement $I(\mathcal{E}_i)$\;
\CommentSty{/* Store promising emitters in emitters buffer */}\\
\If{$I(\mathcal{E}_i) > 0$}{
    $\mathcal{Q}_{\text{Em}} \leftarrow \mathcal{E}_i$\;
}
}
\CommentSty{/* Emitters step */}\\
Calculate pareto fronts in $\mathcal{Q}_{\text{Em}}$\;
\While{$\nicefrac{2}{3}K_{Bud}$ not depleted}{
Sample $\mathcal{E}_i$ from \emph{non-dominated emitters} in $\mathcal{Q}_{\text{Em}}$\;

\While{\textbf{not} $terminate(\mathcal{E}_i)$}{
Generate offspring population $P^m_{\gamma}$ from $P_{\gamma}$\;

\emph{Eval}($\Tilde{\theta}_j$), $\forall \Tilde{\theta}_j \in P^m_{\gamma}$; ~~ \CommentSty{\SLASH\SLASH Evaluate population}\\

$\mathcal{A}_{\text{Rew}} \leftarrow \Tilde{\theta}_j, ~~ \forall \Tilde{\theta}_j \in P^m_{\gamma} \mid r(\Tilde{\theta}_j) > R_{\gamma}$; \CommentSty{\SLASH\SLASH Store high rewarding policies}\\

$\mathcal{Q}^i_{\text{Cand\_Nov}} \leftarrow \Tilde{\theta}_j, ~~ \forall \Tilde{\theta}_j \in P^m_g \mid \eta(\Tilde{\theta}_j) > \eta_i$; \CommentSty{\SLASH\SLASH Store high novelty policies}

Generate $P_{\gamma+1}$ from best $\Tilde{\theta}_j \in P^m_{\gamma} \bigcup P_{\gamma}$\;
Update $I(\mathcal{E}_i)$ and $R_{\gamma}$\;

\If{$terminate(\mathcal{E}_i)$}{
$\mathcal{A}_{\text{Nov}} \leftarrow Sample(\mathcal{Q}^i_{\text{Cand\_Nov}}, N_Q)$; \CommentSty{\SLASH\SLASH Store $N_Q$ novel policies in archive}\\
Discard emitter $\mathcal{E}_i$\;
}
}
}
\end{algorithm}
\subsubsection*{Emitters}
\looseness=-1
Emitters are what \gls{name} uses to optimize the reward.
An emitter in an instance of a \emph{reward-based \gls{ea}}.
While any reward optimization method can be used, in this paper we base the emitters on an \emph{elitist \gls{ea}}, similarly to the work of \cite{paolo2021sparse}.
At each generation, the emitter selects the population among the best performing policies $\Tilde{\theta_j}$ from the previous generation's population and offsprings, while the offsprings themselves are generated according to Eq. \eqref{eq:offsprings}.
Using an elitist \gls{ea} removes the need to estimate a covariance matrix from the emitter population.
This estimation can be unstable in situations in which the population size is lower than the dimensionality of the space, as can be often the case when working with neural networks.
To prevent this instability issue, methods like CMA-ES \citep{hansen2016cma} take into account information about older generations when estimating the covariance.
This can render the estimation of the quality of an emitter from its initial generations less reliable, limiting the performance of a method like \gls{name} which discards less promising emitters according to their initial performance.

\looseness=-1
Each one of the emitters $\mathcal{E}_i$ used by \gls{name} consist of a population $P_{\gamma}$ of size $M_{\mathcal{E}}$ of policies $\Tilde\theta_i \in \Theta$, its offspring population $P^m_{\gamma}$ of size $m\times M_{\mathcal{E}}$, a \emph{novelty candidates buffer} $\mathcal{Q}_{\text{Cand\_Nov}}$ in which the most novel policies are stored, a generation counter $\gamma$, and a tracker for the highest reward found until now $R_{\gamma}$.
At the same time, the emitter also tracks two novelties, $\eta_{\gamma}$, that is the novelty of the most novel policy found until generation $\gamma$, and the \emph{emitter novelty}, $\eta(\mathcal{E}_i)$, corresponding to the novelty of the policy generating the emitter.
The emitter is initialized from the policy $\theta_i$ by sampling the $M_{\mathcal{E}}$ policies in its initial population $P_0$ from the distribution $\mathcal{N}(\theta_i, \sigma_i I)$.
To reduce the overlap of the emitter's search space with the ones of possible nearby emitters, \gls{name} shapes $\mathcal{N}(\theta_i, \sigma_i I)$ such that the distance between $\theta_i$ and the closest $\theta_j$ corresponds to 3 standard deviations.
This is done by initializing $\sigma_i$ as:
\begin{equation}
\sigma_i = \frac{\min_j(\text{dist}(\theta_i, \theta_j))}{3}, ~~ \forall\Tilde\theta_j \in \Gamma_g^m \cup \Tilde\Gamma_g.
    \label{eq:sigma_i}
\end{equation}
During its evaluation, an emitter tracks also its own \emph{emitter improvement} $I(\mathcal{E}_i)$, a metric that is then used by \gls{name} to select which emitters to prioritize and which to discard, allowing a better allocation of evaluation budget.
A positive $I(\mathcal{E}_i)$ means that the emitter can improve on its initial reward.
On the contrary, $I(\mathcal{E}_i) \leq 0$ means that the chances for the emitter to find better rewards are low, so it is not worth allocating more evaluation budget to it.

The improvement is calculated as the difference between the average reward obtained during the most recent and the initial generations of the emitter:
\begin{equation}
 I(\mathcal{E}_i) = \frac{1}{\lambda M_{\mathcal{E}}} \left( \sum_{\gamma=T-\nicefrac{\lambda}{2}}^{T} \sum_{j=0}^{M_{\mathcal{E}}}r_{(\gamma,j)} - \sum_{\gamma=\gamma_0}^{\nicefrac{\lambda}{2}} \sum_{j=0}^{M_{\mathcal{E}}}r_{(\gamma,j)} \right),
    \label{eq:improvement}
\end{equation}
where $T$ is the last evaluated generation, $r(\gamma, j)$ is the reward of policy $\Tilde\theta_j \in P_{\gamma}$ and $\gamma_0$ is the generation at which the emitter is at the beginning of its evaluation.

An emitter is terminated once a \emph{termination criterion} is reached.
There can be many termination criteria, depending on the kind of algorithm used as emitter.
In this work, we use the one introduced by \cite{paolo2021sparse}.
This criterion is directly inspired by the \emph{stagnation criterion} used for the CMA-ES algorithm and introduced by \cite{hansen2016cma} and stops the emitter once there is no more improvement on the reward.
This is calculated by tracking the history of the rewards obtained by the emitter over the last $120 + 20 * n \slash M_{\mathcal{E}}$, where $n$ is the size of the parameter space $\Theta$ and $M_{\mathcal{E}}$ is the population size of the emitter.
The termination condition is met if the \emph{maximum} or the \emph{median} of the last 20 rewards is lower than the \emph{maximum} or the \emph{median} of the first 20 rewards.

\subsubsection*{Bootstrap step}
\looseness=-1
The candidate emitters buffer $\mathcal{Q}_\text{Cand\_Em}$ contains all the rewarding policies found during the exploration phase.
During the bootstrap step, emitters are initialized from these policies starting from the most novel ones with respect to the reward archive $\mathcal{A}_\text{Rew}$.
This allows \gls{name} to focus more on the less explored areas of the rewarding behavior space $\mathcal{B}_\text{Rew}$.

\looseness=-1
Once an emitter $\mathcal{E}_i$ has been initialized, it is executed for $\lambda$ generations to evaluate its potential for improving the reward by calculating its initial \emph{emitter improvement} $I(\mathcal{E}_i)$.
Only the emitters with positive improvement after this initial evaluation phase are added to the \emph{emitter buffer} $\mathcal{Q}_{Em}$ for further evaluation during the emitter step, while the rest are discarded.
This allows \gls{name} to quickly discard emitters whose initializing policy is in a hard-to-optimize local minima of the reward space.
At the same time, it helps in discovering the policies whose behaviors are in the most promising regions of the rewarding behavior space $\mathcal{B}_{\text{Rew}}$.
The whole bootstrap step lasts $\nicefrac{K_{bud}}{3}$ evaluation steps, at the end of which \gls{name} switches to the \emph{emitter step}.

\subsubsection*{Emitter step}
\looseness=-1
During this step, \gls{name} evaluates the emitters that, due to a positive \emph{emitter improvement}, are now present in the \emph{emitter buffer} $\mathcal{Q}_{Em}$.
The step starts by calculating the Pareto front between the improvement $I(\cdot)$ and the emitter novelty $\eta(\cdot)$ of the emitters in the buffer.
The emitter $\mathcal{E}_i$ to run is then sampled from the front of the \emph{non-dominated emitters}.
%Selecting the emitters according to both their novelty and their improvements 
This allows \gls{name} to focus on the most promising and less explored areas of the rewarding search space $\mathcal{B}_{\text{Rew}}$.

\looseness=-1
The policies $\Tilde{\theta_i}$ found during the evaluation of an emitter $\mathcal{E}_i$ can be stored either for their novelty or for the reward they obtain.
At every generation $\gamma$, the policies with a novelty higher than the maximum novelty found by the emitter so far, $\eta_{\gamma-1}$, are stored in the \emph{novelty candidates buffer} $\mathcal{Q}_{\text{Cand\_Nov}}$.
At the same time, the policies with a reward higher than the maximum reward found until $\gamma-1$, $R_{\gamma-1}$, are stored into the \emph{reward archive} $\mathcal{A}_{\text{Rew}}$.
Once these policies have been stored, both $\eta_{\gamma-1}$ and $R_{\gamma-1}$ are updated with the new maximum values.

\looseness=-1
The emitter $\mathcal{E}_i$ is run until either one of these two conditions happens: the $\nicefrac{2}{3}K_{Bud}$ evaluation budget chunk is depleted or a termination condition is met.
The first case leads \gls{name} to update the improvement of $\mathcal{E}_i$ and store it again in the emitters buffer $\mathcal{Q}_{Em}$ for a possible future evaluation.
The algorithm then goes back to the exploration phase.
In the second case, the emitter is terminated and $N_Q$ policies from the emitter's novelty candidate buffer are uniformly sampled to be added to the novelty archive $\mathcal{A}_{\text{Nov}}$.
This allows \gls{name} to save particularly novel solutions to $\mathcal{A}_{\text{Nov}}$ and prevent the search to go back to already explored areas.
Finally, a new emitter to be evaluated is selected from the front of non-dominated emitters.

\looseness=-1

\section{Experiments}
\label{sec:experiments}
\looseness=-1
This section studies how \gls{name} can discover highly rewarding policies while exploring a behaviour representation space learned online.
All of this with minimal previous information about the environment and the task at hand.
\gls{name} will be compared against various baselines.
Moreover, multiple ablation studies will be performed to study which aspects of the method are the most important ones.
In Sec. \ref{sec:stax_exploration} we will evaluate the exploration performance of the algorithms, while the exploitation performance will be studied in Sec. \ref{sec:stax_exploitation}.
An example of the final distribution of the behavior representation learned by the discovered policies is given in Sec. \ref{sec:archive_distr}.
Ablation studies on the factors contributing to the exploration performance and the importance of the \gls{ae} training regime are done respectively in Sec. \ref{sec:stax_ablation} and Sec. \ref{sec:ae_training}.
Finally, in Sec. \ref{sec:learned_bs} we evaluate the quality of the learned \gls{bs}.

\begin{figure}
    \centering
    \includegraphics[width=.5\textwidth]{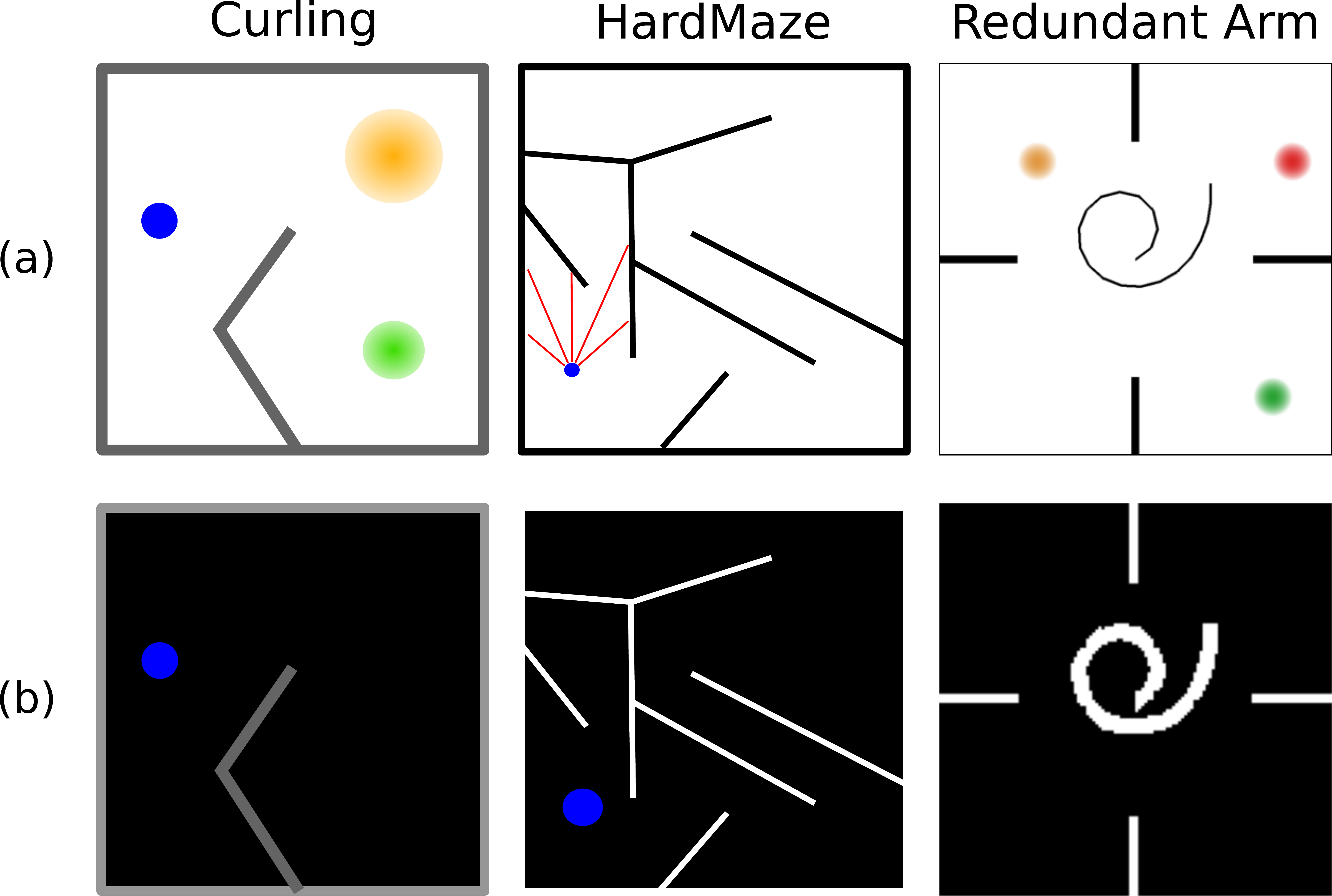}
    \caption{The three testing environments. Row (a) shows the real environments. The reward areas are represented by the green, orange and red circles. Row (b) contains the $64 \times 64$ RGB observations of the environment as seen by the \gls{ae}. The behavior descriptors are generated by sampling 5 of these images along the trajectories.}
    \label{fig:envs}
\end{figure}

In order to perform this analysis, \gls{name} is evaluated on 3 sparse rewards environments, shown in Fig. \ref{fig:envs}.
\setlength{\textfloatsep}{0.3cm}

\looseness=-1
\textbf{Curling}: it consists of a 2 \gls{dof} arm pushing a ball over a table \citep{paolo2021sparse}.
The arm is controlled by a 3 layers \gls{nn} with each layer of size 5.
The input of the controller is a 6-dimensional array containing the $(x,y)$ ball pose and the joints angles and velocities.
The controller outputs a 2-dimensional array containing the speeds of the two joints at the next time-step.
Each policy is run in the environment for 500 timesteps.
The reward is given only if the ball is in one of the two rewarding areas and is higher the closer it is to the center of the area.
The ground truth behavior descriptor used by methods that do not learn the representation is the final $(x,y)$ position of the ball.
The environment, together with the $64 \times 64$ RGB image the \gls{ae} sees during the algorithm execution, is shown in Fig. \ref{fig:envs}.

\looseness=-1
\textbf{HardMaze}: it consists of a 2-wheeled robot whose goal is to navigate a maze with the aid of 5 distance sensors \citep{lehman2008ns}.
The robot, in blue in Fig. \ref{fig:envs}, is controlled by a 2-layers \gls{nn} with each layer of size 5.
The controller receives as inputs the reading of the 5 distance sensors, shown in red in Fig. \ref{fig:envs}, and outputs the speed of the wheels for the next timestep.
The agent receives a reward if the robot reaches one of the 2 reward areas, with the reward being higher the closer to the center the robot stops.
Each policy is run in the environment for 2000 timesteps.
The ground truth behavior descriptor used by methods that do not learn the representation is the final $(x,y)$ position of the robot.
The environment, together with the $64 \times 64$ RGB image the \gls{ae} sees during the algorithm execution, is shown in Fig. \ref{fig:envs}.

\looseness=-1
\textbf{Redundant Arm}: it consists of a 20-DoF arm moving on a 2-dimensional plane \citep{loviken2017online}.
The arm is controlled by a \gls{nn} with 2 layers, each one of size 5.
The input of the controller is the 20-dimensional vector of each joint's positions, while the output consists of the 20-dimensional joints' torque vector.
The policies are run for 100 timesteps each, or until the arm collides either with the wall or itself.
The ground truth behavior descriptor used by methods that do not learn the representation is the final $(x,y)$ position of the end effector.
The reward is given if the end effector reaches one of the three highlighted areas, with the reward being higher the closer the effector is to the center of the reward area.
The environment, together with the $64 \times 64$ RGB image the \gls{ae} sees during the algorithm execution, is shown in Fig. \ref{fig:envs}.

\looseness=-1
In all these environments, \gls{name} builds the behavior descriptors by stacking the low-dimensional representations extracted by the \gls{ae} from multiple high-dimensional observations. 
To this end, 5 samples collected at regular intervals along the trajectories are used during the experiments.

\subsection*{Baselines}
\looseness=-1
\gls{name} is compared against the following baselines:
\begin{itemize}
    \item \textbf{\gls{ns}} \citep{lehman2008ns}: vanilla \gls{ns}, that performs pure exploration in the ground-truth behavior space and does not attempt to improve on the reward;
    \item \textbf{\gls{me}} \citep{mouret2015illuminating}: vanilla MAP-Elites that uses a $50 \times 50$ grid to cover the ground-truth behavior space of each environment;
    \item \textbf{\gls{moo}-NR} \citep{deb2002fast}: a multi-objective evolutionary algorithm optimizing both the novelty in the ground-truth behavior space and the reward of the policies;
    \item \textbf{\gls{taxons}} \citep{paolo2020unsupervised}: that performs pure exploration by learning the behavior descriptor through an \gls{ae} trained during the search process;
    \item \textbf{\gls{serene}} \citep{paolo2021sparse}: that performs exploration through \gls{ns} in the ground-truth behavior space, exploiting any discovered reward through emitters.
\end{itemize}
\looseness=-1
Note that among the baselines, only \gls{taxons} learns the behavior descriptor similarly to \gls{name}.
The other baselines all use the ground-truth behavior descriptor.

\looseness=-1
For each experiment, the given evaluation budget is $Bud = 500000$, with a chunk size of $K_{Bud}=100$.
The population has a size of $M=100$ and each policy generates $m=2$ offsprings.
This is done by using a mutation parameter of $\sigma =0.5$.
At each generation, the number of policies sampled to be added to the novelty archive is $N_Q=5$.
The emitters have a population size of $M_{\mathcal{E}} = 6$ with a bootstrap phase of $\lambda = 6$.
For every experiment, the policies' parameters are bounded in the $[-5, 5]$ range.
All approaches using an \gls{ae} to represent the behavior descriptor use the same structure.
The \gls{ae} consists of an encoder $E(\cdot)$ with 4 convolutional layers of sizes [32, 64, 32, 16], followed by a linear layer projecting the 256-dimensional vector returned by the last convolutional layer into the 10-dimensional feature space.
Each convolutional operation has a kernel of size 4, with a stride of 2 and a padding of 1.
Every layer is followed by a SeLU activation function \citep{Klambauer2017selu}, allowing the self-normalization of the \gls{nn}.
On the contrary, the decoder $D(\cdot)$ starts with a linear layer projecting the 10-dimensional feature vector into a 256-dimensional vector. 
Then it is followed by 4 convolutional layers of sizes [32, 64, 32, 3], each one using a kernel of size 4, a stride of 2, and a padding of 1.
Every layer uses a SeLU activation function, with the exception of the last convolutional one using a ReLU, in order to force the non-negativity of the output value.
The weights of the \gls{ae} are randomly initialized through the default Pytorch initialization.
This is done by sampling the weights of each layer from an uniform distribution $U(-\frac{1}{\sqrt{\omega}}, \frac{1}{\sqrt{\omega}})$, with $\omega$ being the number of learned parameters in the layer.
% This info comes from here: https://discuss.pytorch.org/t/what-is-the-default-initialization-of-a-conv2d-layer-and-linear-layer/16055
The training is done with the Adam optimizer \citep{kingma2014adam} with a learning rate of 0.001.
The results are computed over 15 runs for each experiment and their statistical significance is evaluated by performing a Mann-Whitney test \citep{mann1947test} with Helm-Bonferroni correction \citep{holm1979simple}.
Finally, in each plot, the performances of methods using the ground-truth \gls{bs} are represented with dashed lines, while the methods learning the behavior space are shown through a continuous line.
The code repository is available at: \texttt{https://github.com/GPaolo/STAX.git}.

\section{Results}
\label{sec:stax_results}

\setlength{\textfloatsep}{0.3cm}
\looseness=-1
In this section, the results obtained from the experiments are discussed.
The significance of the results is evaluated through the non-parametric Mann-Whitney U test \citep{mann1947test} with Holm-Bonferroni correction \citep{holm1979simple}.
%SD: not sure that it is necessary to repeat it

\subsection{Exploration}
\label{sec:stax_exploration}
\begin{figure}
    \centering
    \includegraphics[width=.9\textwidth]{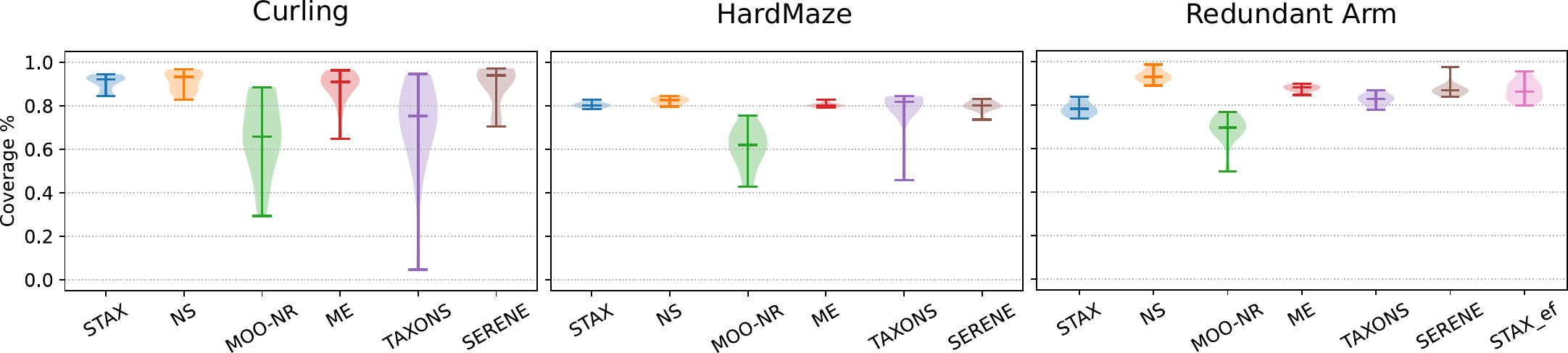}
    \vspace*{-3mm}
    \caption[STAX coverage results]{Final coverage reached by \gls{name} against the different baselines after $5\times 10^5$ evaluations.
    The medians and the extrema are highlighted. The plots are calculated over 15 seeds.}
    \label{fig:stax_cvg_baselines}
\end{figure}

\looseness=-1
This section studies how well \gls{name} can explore in situations of sparse rewards while having minimal information about the environment and the task.
This is done by measuring the \emph{coverage metric} obtained in the \emph{ground truth} \gls{bs} defined in Sec. \ref{sec:experiments} for each one of the tested environments.
The coverage metric is evaluated by dividing said ground truth space into a $50 \times 50$ grid and calculating the percentage of cells occupied during the search.
A cell is considered occupied if a policy reaches it at the end of its evaluation.
Note that, while the coverage is calculated in the ground-truth space, \gls{name} has no access to this space at search time.
The algorithm has to learn a representation from a collection of high-dimensional observations in order to perform the exploration.
This means that the method can also explore areas of the space that are not considered by the coverage metric in the ground-truth space.
An example of this is the Curling environment, in which a single final position of the ball - the one considered in the ground-truth \gls{bs} - can correspond to multiple arm positions that are represented by \gls{name}.
At the same time, the strongest baseline with respect to this metric is \gls{ns} which has direct access to the space in which the coverage is calculated, providing an upper-bound value for our experiments.

\looseness=-1
Fig. \ref{fig:stax_cvg_baselines} shows the coverage reached by our method and all the tested baselines.
It can be seen that on the Curling environment \gls{name} performs similarly to \gls{ns}, with a mean final coverage of $90.8 \%$ for \gls{name} compared to the $91\%$ for \gls{ns} ($p=.77$).
In the other two environments, \gls{name} reaches lower coverage compared to \gls{ns}.
The difference is small on the HardMaze, $80.3\%$ for \gls{name} versus $82.2\%$ for \gls{ns} ($p=1.5\times 10^{-2}$), but it is higher on the Redundant Arm environment with $78.2\%$ for \gls{name} against the $93.3\%$ obtained by \gls{ns} ($p = 7.31 \times 10^{-5})$.

\looseness=-1
The reason for \gls{name}'s low performances on this last environment are due to \gls{name} learning to represent the whole arm configuration rather than only the end effector position, thus maximizing diversity in dimensions not considered by the coverage metric.
On the contrary, the $86.6\%$ of coverage reached by \gls{name} when the \gls{ae} is only shown the end effector position, rather than the whole arm, STAX\_ef in the Redundant Arm plot in Fig. \ref{fig:stax_cvg_baselines}, are comparable to the coverage of $87\%$ reached by \gls{serene} ($p=.47$).
The other methods using the hand-designed ground-truth \gls{bs} to drive the search - \gls{me} and \gls{serene} - reach high levels of coverage comparable to \gls{ns} on Curling (\gls{me}: $p=.77$, \gls{serene}: $p=.77$), but slightly lower on both Redundant Arm (\gls{me}: $p=1.7 \times 10^{-4}$, \gls{serene}: $p=5.8 \times 10^{-4}$) and HardMaze (\gls{me}: $p=1.7 \times 10^{-2}$, \gls{serene}: $p=2.6 \times 10^{-2}$).
This is expected given that both methods perform the search in the same space in which the coverage metric is computed but also optimize the reward.
The good performance of \gls{name} is instead obtained with minimal information about the task and the space in which information is gathered.
At the same time, \gls{moo}-NR struggles in all environments, likely because once a rewarding solution is found, it will dominate all the non-rewarding solutions, strongly limiting the exploration of the method.

\looseness=-1
\gls{taxons} also obtains high coverage, with the notable exception of the Curling environment.
The culprit of this loss of performance is likely the presence of the 2-DOF arm in the image fed to the \gls{ae}, as shown in Fig. \ref{fig:envs}, that can act as a distractor in situations in which only the final position of the ball is interesting.
At the same time, the presence of the arm is not a hindrance to the performances of \gls{name}.
This is likely due to both the higher amount of data on which the \gls{ae} is trained - the 5 frames sampled along the trajectory for \gls{name} compared to only the last frame for \gls{taxons} - and the better selection of new policies according to the \gls{moo} based approach, performed by \gls{name}.
The effects of these factors on the performance of \gls{name} will be studied in Sec. \ref{sec:stax_ablation}.

\subsection{Exploitation}
\label{sec:stax_exploitation}
\looseness=-1
\begin{figure}[!t]
    \centering
    \includegraphics[width=\textwidth]{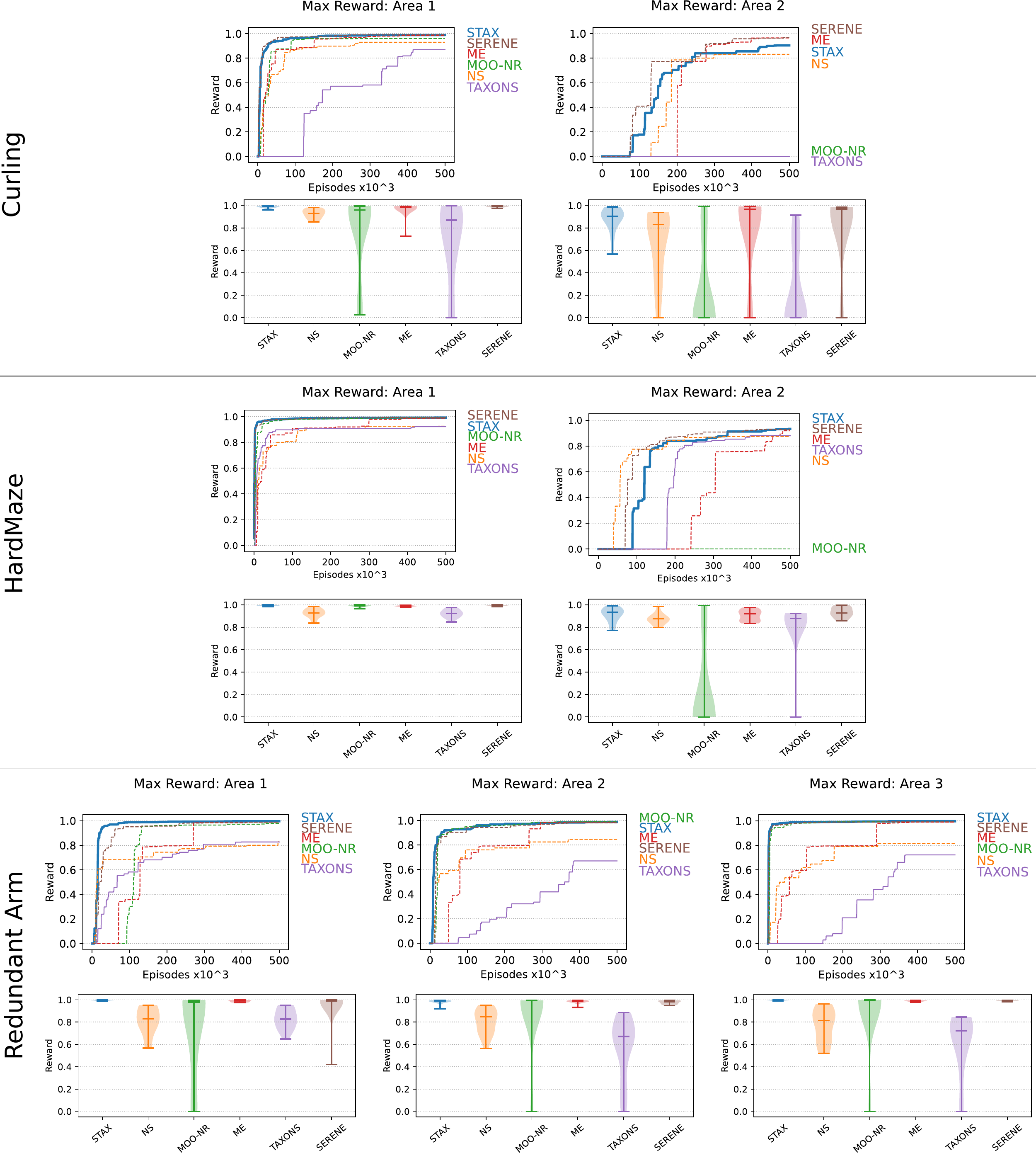}
    \vspace*{-3mm}
    \caption[STAX reward results]{Maximum reward reached in all the reward areas by \gls{name} against the different baselines.
    For each environment, the top row represents the median maximum reward with respect to the whole evaluation budget. The bottom row represents the final maximum reward obtained by the algorithms. The medians and the extrema are highlighted. 
    All plots have been calculated over 15 seeds.}
    \label{fig:stax_cvg_reward}
\end{figure}
The maximum reward achieved by the algorithms in all the reward areas is shown in Fig. \ref{fig:stax_cvg_reward}.
Using emitters to exploit the reward allows \gls{name} to reach high rewards in a few evaluations. 
These performances are similar to the ones obtained by \gls{serene} on Curling ($p=.53$) and HardMaze ($p=.71$) and slightly higher on Redundant Arm ($p=1.4\times 10^{-2}$), thanks to the fact that the reward exploitation performed by the emitters does not rely on any behavior descriptor.
Among the other baselines performing reward improvement, the best performing one is \gls{me}, capable of reaching high values on all reward areas, but at a much slower pace than \gls{name}.
This is not the case for the multi-objective approach \gls{moo}-NR, which can always find at least one of the multiple reward areas, but then tends to extensively focus on it, instead of also exploring other areas.
For this reason, only the easiest reward area is exploited to high values in all environments, while the harder reward area is seldom exploited.
On the contrary, while \gls{ns} and \gls{taxons} can perform good exploration, they cannot reach high reward levels very quickly, with \gls{taxons} being consistently worse in this regard than any other algorithm ($p = 2.5 \times 10^{-4}$).
This is due to the lack of any reward-exploitation mechanism present in both methods.
This is even more noticeable in the redundant arm environment, where even if \gls{taxons} can reach higher coverage levels than \gls{name} ($p=6.9 \times 10^{-3}$), the absence of any reward improving mechanism leads to very low performances on all reward areas.

\subsection{Final archives distribution}
\label{sec:archive_distr}
\begin{figure}
    \centering
    \includegraphics[height=0.9\textheight]{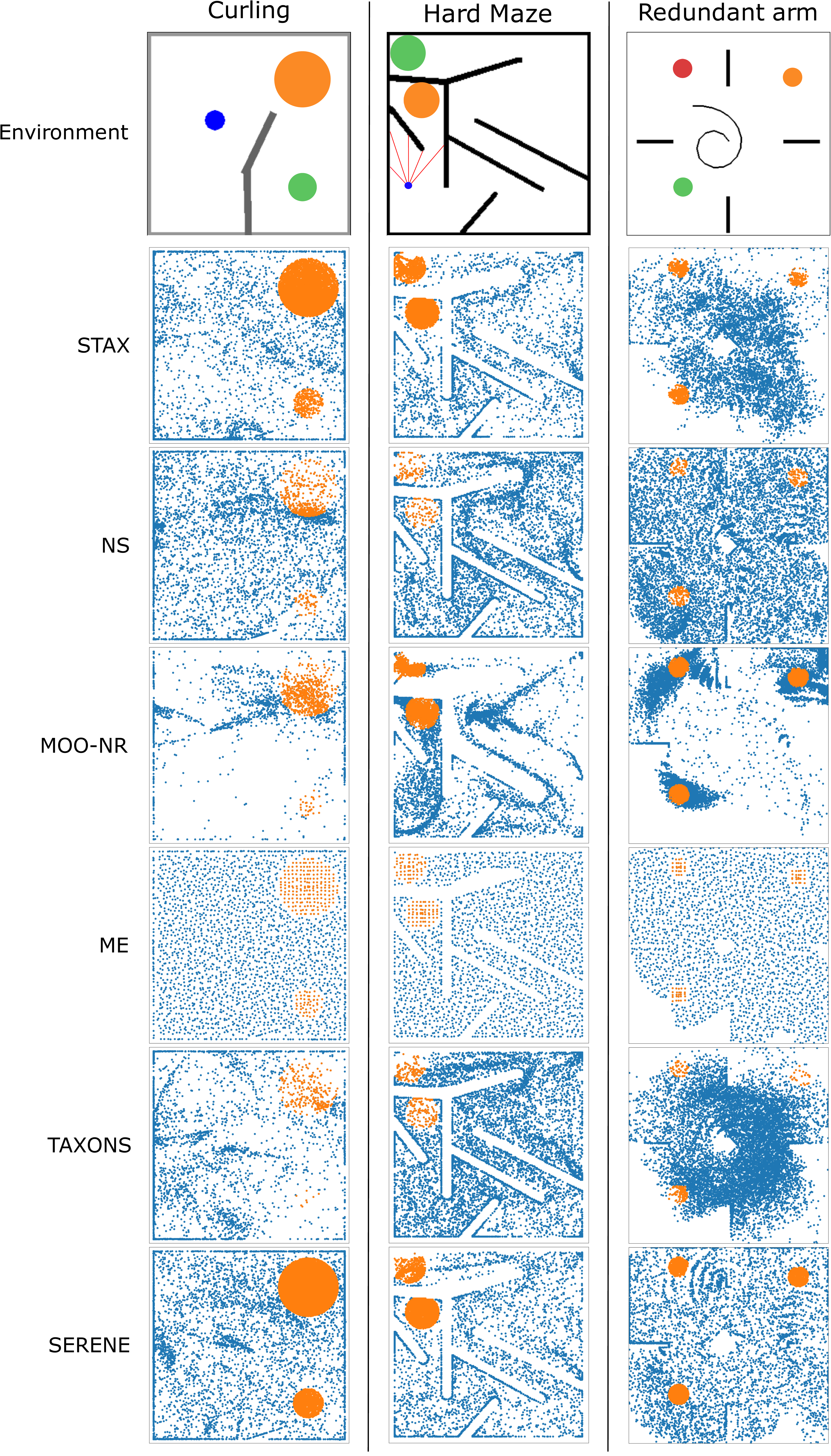}
  \vspace{-3mm}
    \caption{Distribution of the behavior descriptors of the archived policies. On each column are shown the results for an environment, while on each row is shown the distribution for each experiment. The archive plotted are from the runs achieving the highest coverage. In blue are the policies outside of the reward area, while in orange are the policies within the reward area.}
    \label{fig:stax_final_arch}
\end{figure}
\looseness=-1
An example of the final distribution of the behaviors representations for the policies in the final archives is shown in Fig. \ref{fig:stax_final_arch}.
Each point represents a policy.
In blue are shown the policies present in the novelty archive $\mathcal{A}_{\text{Nov}}$, while in orange are the policies in the reward archive $\mathcal{A}_{\text{Rew}}$.
For the baselines not using the double archives structure, the blue points represent the policies that did not receive any reward, considered \emph{exploratory}, while the orange points represent the rewarding policies.
If a method is capable of properly exploring the behavior space, the blue dots should cover as much as possible of the space.
At the same time, a method capable of optimizing the reward, should be able to focus on the reward areas, thus producing many solutions reaching said areas (orange dots).

\looseness=-1
From the figure, it is possible to see how emitter-based methods, \gls{name} and \gls{serene}, densely cover the reward areas discovered during exploration, while \gls{ns} and \gls{taxons} do not have this effect due to the lack of any exploitation mechanism.
At the same time, the row of \gls{moo}-NR shows how once reward areas are discovered, the method mainly focuses on those.
Finally, the figure shows how \gls{me} very uniformly covers the search space compared to the other methods, thanks to the discretization of the behavior space. 

\subsection{Exploration ablation studies}
\label{sec:stax_ablation}
\looseness=-1
This section studies the contributing factors to the exploration results obtained by \gls{name}.
The study focuses on two aspects of the algorithm: the multi-objective approach for policy selection and the multiple observations used to generate the behavior descriptor of a policy.
Four ablated variants of \gls{name} are considered:
\begin{itemize}
\item \textbf{\gls{name}\_multi}: it is the vanilla version of \gls{name}. It uses both the multi-objective policy selection between novelty and surprise and the 5 observations sampled along the policy trajectory to generate the behavior descriptor;
\item \textbf{\gls{name}\_single}: this variant still uses the multi-objective policy selection strategy, but the behavior descriptor is calculated only from the last observation. This baseline is used to evaluate how important is to use points along the whole trajectory rather than just the last one;
\item \textbf{\gls{name}-ALT\_multi}: this variant uses the same strategy used by \gls{taxons} to select between novelty and surprise, sampling either one of the two at each generation. The behavior descriptor is generated by using 5 observations sampled at regular intervals along the trajectory. This baseline is used to evaluate the importance of the new policy-selection strategy used by \gls{name};
\item \textbf{\gls{name}-ALT\_single}: as the previous variant, here the \gls{taxons} policy selection strategy is used. Moreover, the behavior descriptor is generated by only the last observation of the trajectory.
\end{itemize}
Both the coverage and the maximum reward reached by each variant over each reward area are analyzed.

\begin{figure}[!th]
    \centering
    \includegraphics[width=\textwidth]{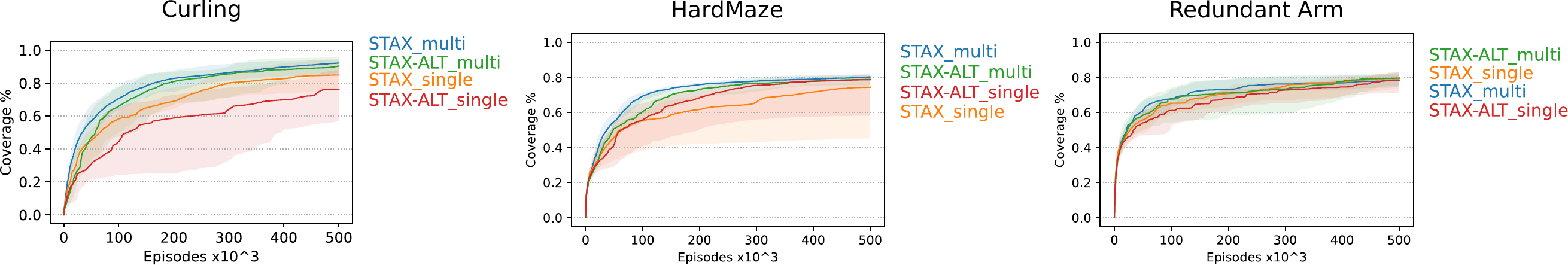}
    \vspace*{-3mm}
    \caption[STAX ablation experiments - coverage results]{Median coverage with respect to the given evaluation budget reached by \gls{name} against the ablated versions of the algorithm. The shaded areas represent the 10 and 90 percentile calculated over 15 seeds.}
    \label{fig:stax_ablation_cvg}
\end{figure}

\begin{figure}[!h]
    \centering
    \includegraphics[width=.8\textwidth]{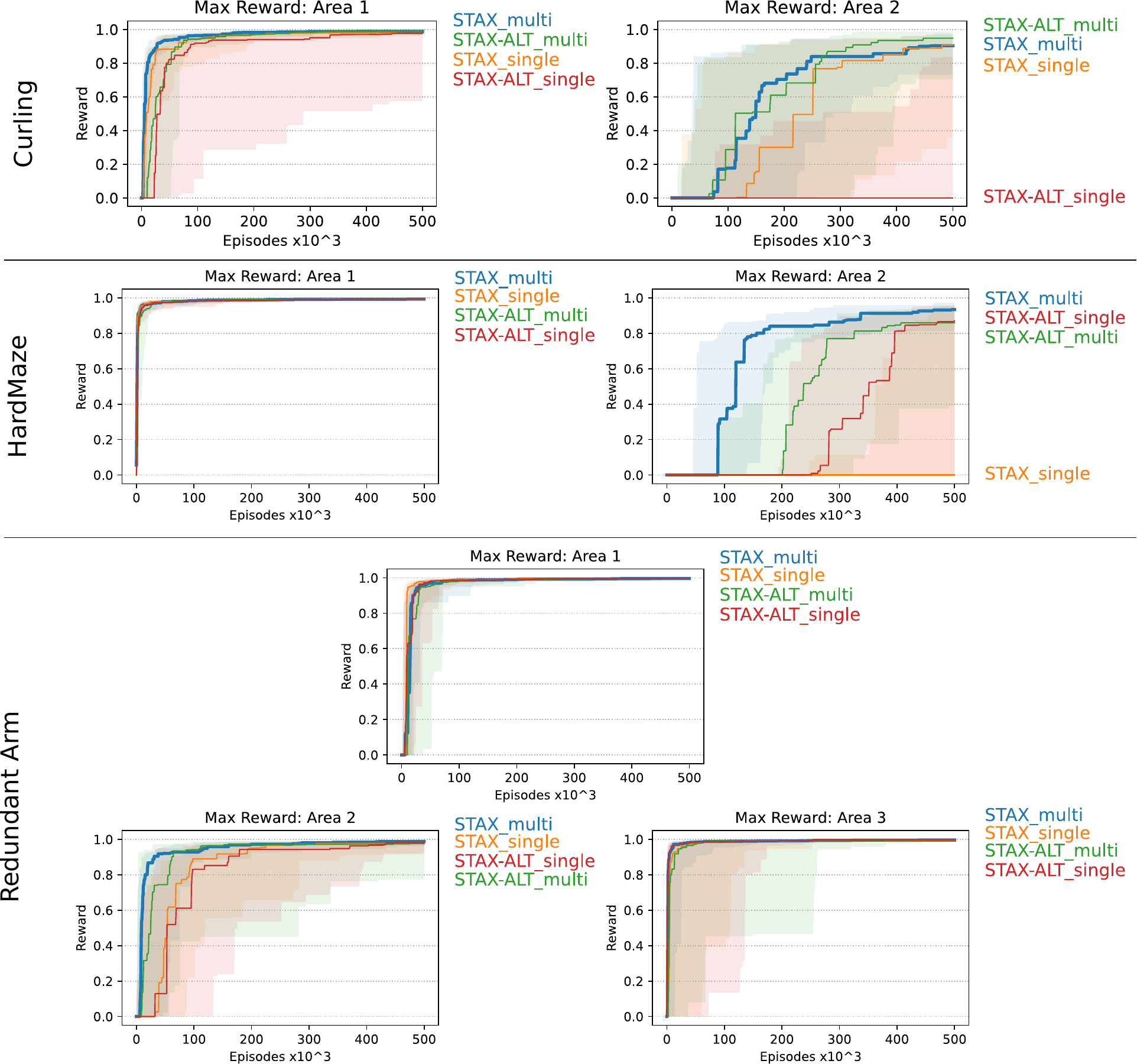}
    \vspace*{-3mm}
    \caption[STAX ablation experiments - reward results]{Median maximum reward reached in all the reward areas by \gls{name} against the ablated versions of the algorithm. The shaded areas represent the 10 and 90 percentile calculated over 15 seeds.}
    \label{fig:stax_ablation_rew}
\end{figure}

\looseness=-1
The average coverage is shown in Fig. \ref{fig:stax_ablation_cvg}.
It is possible to see that the final reached coverage is similar for all variants on all environment ($p > .05$), with the exception of the Curling environment in which the variants using multiple observations reach higher final coverage compared to both \gls{name}\_single ($p < 0.028$) and \gls{name}-ALT\_single ($p <  4.33 \times 10^{-5}$).
Nonetheless, the variants using multiple observations reach higher coverage earlier during the runs compared to the single observation variants.
This is likely due to the \glspl{ae} of these variants being trained on 5 times more data than the ones of the variants using a single observation. 
Moreover, being the data collected along the whole trajectory, it provides a more diverse collection of data from the observation space, making it easier for the algorithm to learn a good representation.

\looseness=-1
The improved performance provided by using multiple observations can be seen also when analyzing the maximum reward reached in the environments, as shown in Fig. \ref{fig:stax_ablation_rew}.
While the final reward reached for the different reward areas is similar for all the methods, \gls{name}\_multi tends to reach high rewards earlier in the runs compared to the ablated approaches.
% Moreover, the harder to reach a reward area is, the more \gls{name}\_multi tends to outperform the ablated versions, as it can be seen for the Reward Area 2 of the HardMaze environment

% The biggest difference between the methods' performance are noticeable when reward areas are harder to reach, as is the case for the Reward Area 2 for both the Curling and the HardMaze environments.
% In these situations, the hard

% In these settings, the versions using multiple observations to build the behavior descriptor reach higher rewards compared to the ones using only the last observation 

% for the hard to reach reward areas present in Curling

% The main advantage of the full method can be seen for the reward area 2 for both the Curling and the HardMaze environments.

% In each of the reward areas of all environments, \gls{name}\_multi reaches the highest performances earlier than the other methods.
% In general, the methods using only the last observation to extract a description of the behavior of a policy perform the worst.
% At the same time, the multi-objective policy selection method, used by both \gls{name}\_multi and \gls{name}\_single, has a weaker but non negligible effect on both exploration and the exploitation.
% It can be seen in fact that the version using both multiple observations and the multi-objective policy selection strategy performs consistently better than all the other variants.

\subsection{Autoencoder training regime}
\label{sec:stax_learned_bs}
\looseness=-1
This section analyzes how the way \gls{bs} is learned through the \gls{ae} influences the search.
In this regard, the study focuses on two aspects: how important it is to learn the representation versus just using a random one and if retraining from scratch the \gls{ae} at each training episode has any influence on the search process.
In \gls{name} the \gls{ae} is continuously trained across different training episodes. 
This means that similarly to what is done by \cite{paolo2020unsupervised}, the training of the \gls{ae} is resumed at every training episode.
This produces a \emph{curriculum effect} over the borders of the explored space due to the training on the last generation of the population and offsprings.
The curriculum effect is also given by training the \gls{ae} over the archives, even if this contribution is small at the beginning of the search when the archives contain only a few elements.

\begin{figure}[!h]
    \centering
    \includegraphics[width=.95\textwidth]{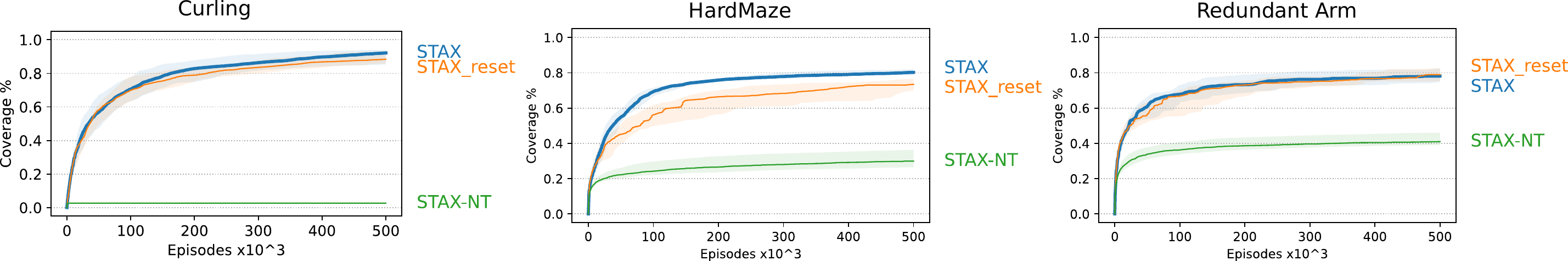}
    \vspace*{-3mm}
    \caption[STAX learned BS experiments - coverage results]{Median coverage with respect to the given evaluation budget reached by \gls{name} against the other versions of the algorithm. The shaded areas represent the 10 and 90 percentile calculated over 15 seeds.}
    \label{fig:stax_reset_cvg}
\end{figure}

\begin{figure}[!h]
    \centering
    \includegraphics[width=.75\textwidth]{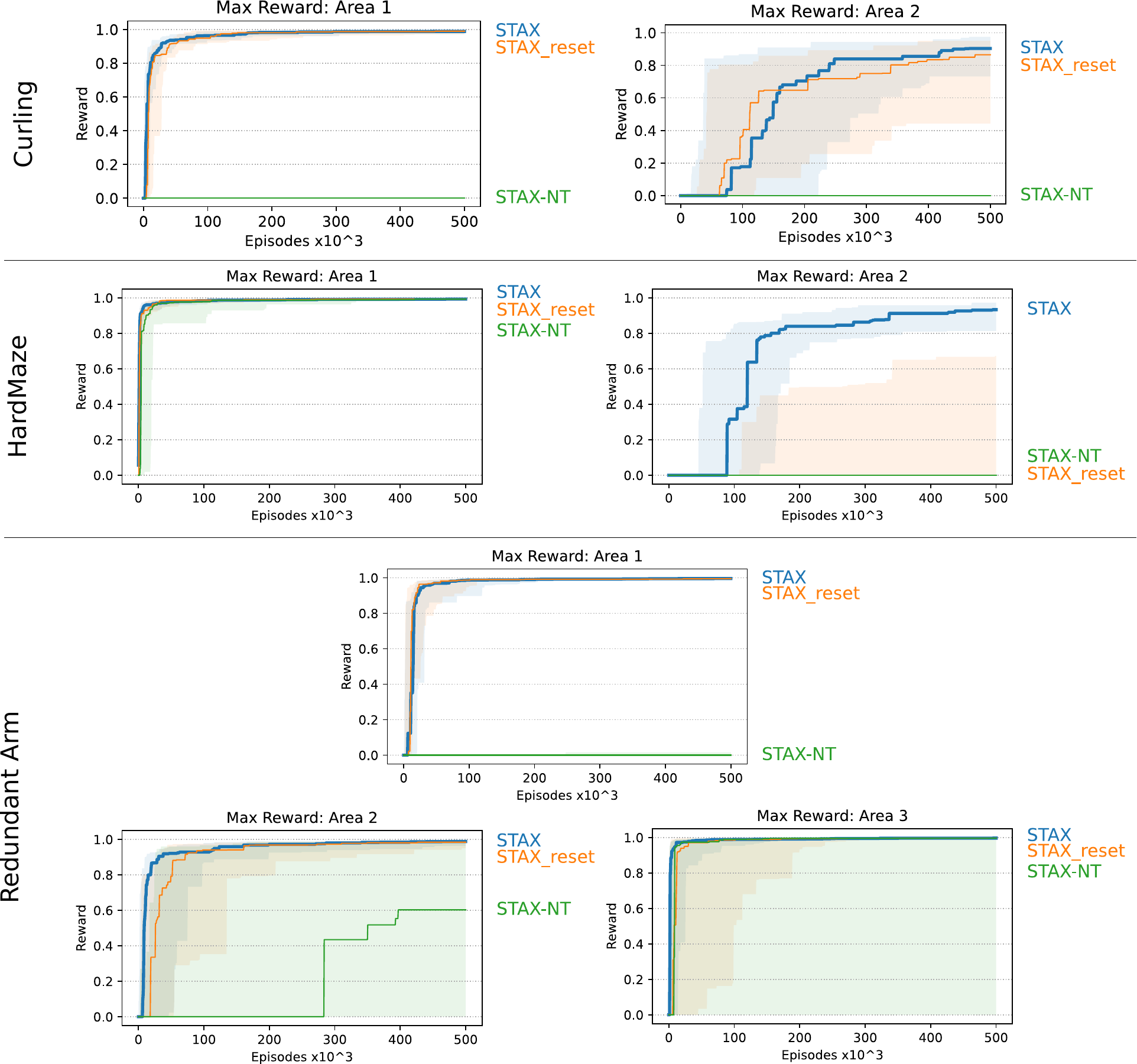}
    \vspace*{-3mm}
    \caption[STAX learned BS experiments - reward results]{Median maximum reward reached in all the reward areas by \gls{name} against the other versions of the algorithm. The shaded areas represent the 10 and 90 percentile calculated over 15 seeds.}
    \label{fig:stax_reset_rew}
\end{figure}

\looseness=-1
To analyze these two aspects, \gls{name} is compared against 2 variants:
\begin{itemize}
    \item \textbf{\gls{name}-NT}: in which the search is driven through an \gls{ae} whose weights are randomly sampled at the beginning of the search and not modified anymore;
    \item \textbf{\gls{name}\_reset}: in which the weights of the \gls{ae} are randomly resampled before each training episode. This means that the \gls{ae} is retrained from scratch at every training episode.
    This effectively removes any memory from previous iterations from the \gls{ae}. 
\end{itemize}
Thanks to the first variant, it is possible to analyze if a random representation is enough to push for exploration and how important is the autonomous learning of \gls{bs}.
The last variant allows studying the importance of the curriculum effect given by the continuous training of the \gls{ae} versus the one provided by the data collected in the archive.
Note that the only change among all these versions of \gls{name} is the \gls{ae} training regime. 
The behavior descriptor is still generated as described in Sec. \ref{sec:stax_exploration}.
The coverage results for the 3 tested environments are shown in Fig. \ref{fig:stax_reset_cvg}, while the rewards reached in each reward area are shown in Fig. \ref{fig:stax_reset_rew}. 

Not surprisingly, the results show that training the \gls{ae}, rather than using a randomly generated one, greatly helps the exploration process. 
The random representations are not enough to discover all the areas of the ground-truth \gls{bs}: 
\gls{name}-NT has significantly lower coverage on all the tested environments compared to the versions in which the \gls{ae} is trained ($p< 3.5 \times 10^{-6}$).
% MAYBE REMOVE
The effect is extreme in the Curling environment in which, to obtain good exploration, it is not enough to randomly move the arm, but it is necessary to properly hit the ball.
In the HardMaze and the Redundant Arm environments, the non-trained versions can explore the easier-to-reach areas of the space, but not reach high levels of coverage.

\looseness=-1
This is reflected in the reward obtained by the methods, shown in Fig. \ref{fig:stax_reset_rew}.
In Curling, random exploration is not enough to discover rewards due to the complex interaction between the ball and the arm, leading to \gls{name}-NT not being able to obtain rewards.
At the same time, the random representation suffices to explore just enough to discover the easy-to-reach reward areas in the easier dynamics of the HardMaze and Redundant Arm, allowing the emitters to exploit them.
%%%

\looseness=-1
On the contrary, the continuous training of the \gls{ae} has a small effect on the coverage:
\gls{name} performs similarly to \gls{name}\_reset on both Redundant Arm ($p=0.27$) and only slightly better on Curling ($p=0.038$) and HardMaze ($p=8.18 \times 10^{-6}$).
This means that the archive can provide enough of a curriculum when learning a representation of \gls{bs}.
The rewards obtained by the two methods are also similar for all environments on all reward areas, with the exception of the harder-to-reach reward area in the HardMaze environment, for which \gls{name} reaches much higher rewards than \gls{name}\_reset ($p = 4.49 \times 10^{-5}$).
The high difference in reward here is due to the fact that this reward area is in the farthest zone from the starting position of the robot.
This means that the small difference in exploration between the two methods often prevents \gls{name}\_reset to discover it and thus to exploit it.
% This hints to the fact that even though the archive can provide enough of a curriculum while learning the representation, in some situation this curriculum effect can be not strong enough. 

% \looseness=-1
% Fig. \ref{fig:stax_reset_rew} shows the rewards the three versions of \gls{name} obtain on the different reward areas. 
% It is possible to see that for the Curling environment, where 

% These experiments clearly show that each environment has different dynamics when it comes to exploration.
% This strengthens our assumption that hand-designing a \gls{bs} in order to properly explore can be difficult and require adaptations to each single situation.
% For this reason, it is important to design algorithms like \gls{name} that can learn said \gls{bs} online while starting with minimal prior information.
% These algorithms should adapt to all environment dynamics by taking advantage as much as possible of the data generated during the search.

\subsection{Learned behavior space}
\label{sec:learned_bs}
\begin{figure}[!t]
    \centering
    \includegraphics[width=0.7\textwidth]{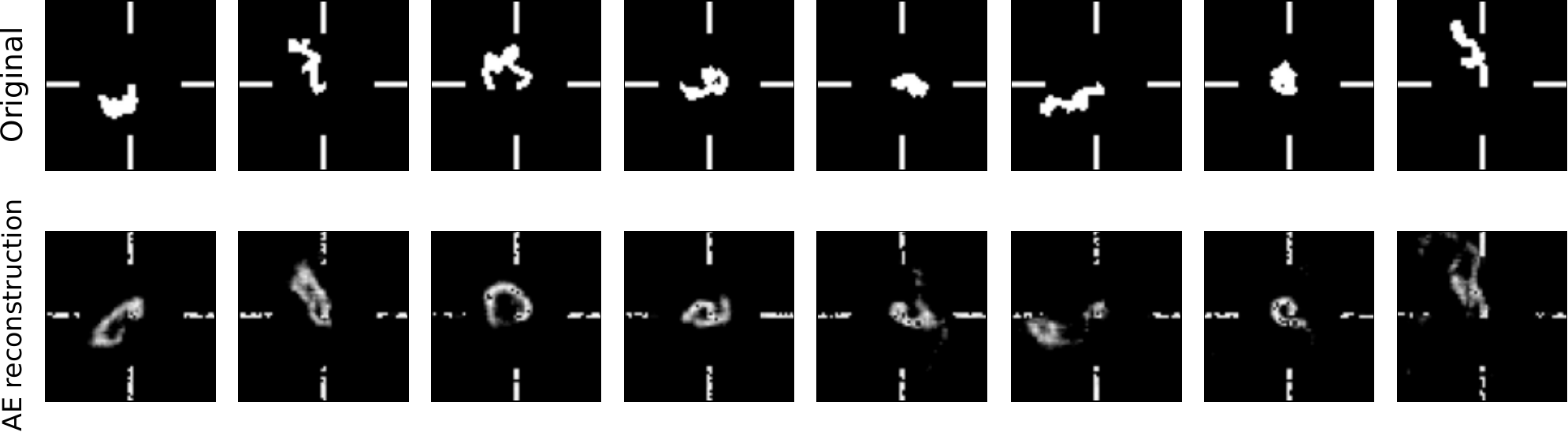}
    \vspace*{-3mm}
    \caption[STAX AE reconstruction]{Reconstruction of the \gls{ae} trained during the search performed by \gls{name}. The first row shows the original $64 \times 64 \times 3$ images. The second row shows the reconstructions of the images produced by the \gls{ae}.}
    \label{fig:stax_ae_reconstruction}
\end{figure}

\looseness=-1
This section studies how well the trained \gls{ae} can represent the behavior space and how close the learned representation is to the ground truth one.
In Fig. \ref{fig:stax_ae_reconstruction} are shown some $64\times64$ observations collected during the evaluation of policies on the Redundant Arm environment (top row) with the respective \gls{ae} reconstructions (bottom row).
This environment provides the hardest to reconstruct observations, given the presence of the whole arm in the images.
It is possible to see from the figure that the reconstructed image is not perfect, even though the position of the arm is clear.
Nonetheless, this level of reconstruction seems to be enough to push for good exploration in the environment, as seen in Sec. \ref{sec:stax_exploration}.

\begin{figure}[!t]
    \centering
    \includegraphics[width=0.65\textwidth]{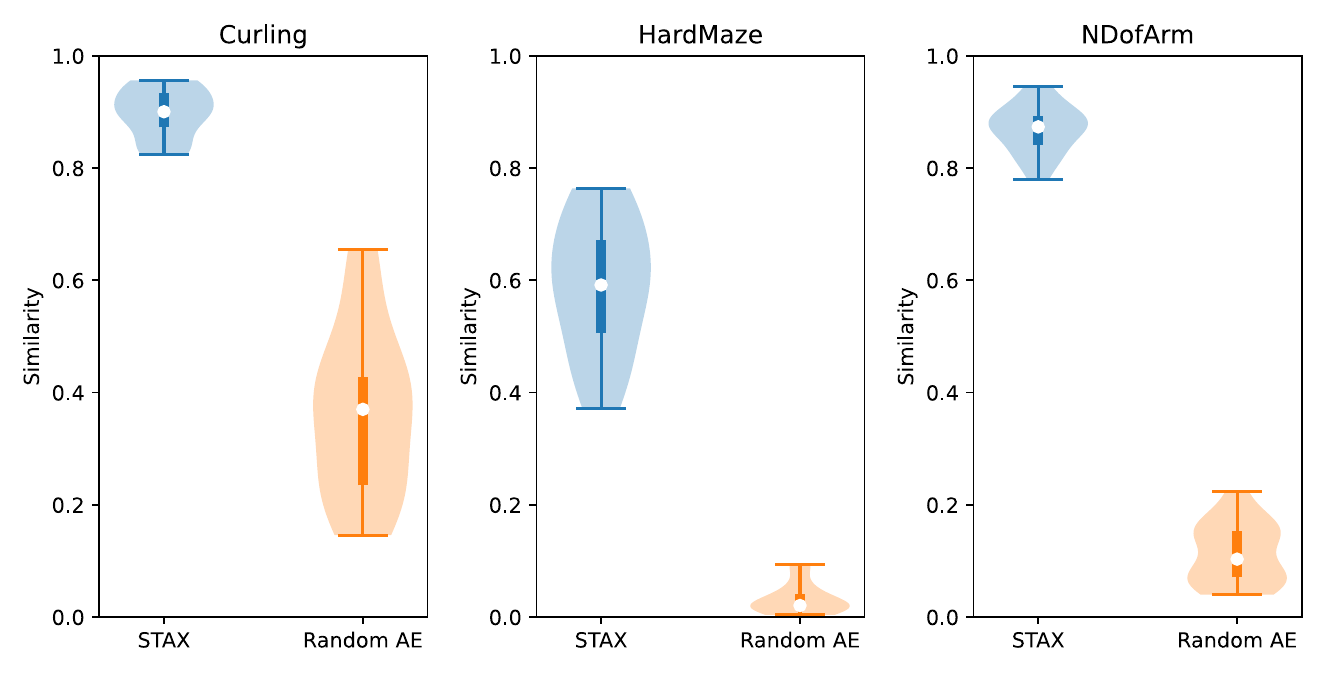}
    \vspace*{-3mm}
    \caption[STAX BD similarity]{Similarity between the ground truth behavior descriptor and the one provided by the \gls{ae} for the three environments. The violins are calculated over 15 seeds.}
    \label{fig:stax_bd_similarity}
   	\vspace{-1mm}
\end{figure}

\looseness=-1
To give a quantitative estimate of the similarity between the learned representation and the ground truth one, we calculated the similarity between the correlation matrices of the ground truth behavior descriptors and the learned descriptors of the policies in the final collections. 
This is done through the following formula \citep{herdin2005correlation}:
\begin{equation}
    s(C_1, C_2) = 1 - \frac{tr(C_1 \cdot C_2)}{||C_1|| \cdot ||C_2||}, 
\end{equation}
where $C_1$ and $C_2$ are the two correlation matrices and the norm is the Frobenius norm.
The metric varies between [0,1] and allows estimating how meaningful is a representation compared to the other.
The higher the value of the metric, the closer we can consider the two representations.
For comparison, we also evaluate $s(\cdot, \cdot)$ between the correlation matrix of the ground truth descriptor and the representation given by a random \gls{ae} over the same observations.
The results are shown in Fig. \ref{fig:stax_bd_similarity}.
\begin{figure}[!h]
	\centering
	\includegraphics[width=0.5\textwidth]{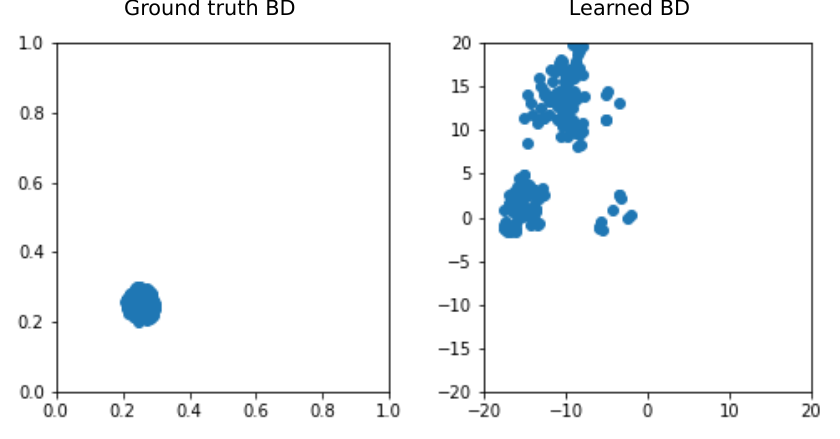}
	\vspace*{-3mm}
	\caption{Ground truth descriptor for the policies in reward area 0 (left) compared to the first 2 principal components of the learned descriptor for the same policies (right). The learned descriptor is divided into multiple areas due to the stacking of frames and the learning of the representation by the \gls{ae}.}
	\label{fig:stax_split_representation}
	\vspace{-3mm}
\end{figure}

\looseness=-1
It is possible to see how representation leaned by \gls{name} reaches high values of similarity compared to the ground truth descriptor on all environments $(p < 4.23 \times 10^{-5})$.
Moreover, the figure also shows that the representation provided by a random \gls{ae} is much less meaningful with respect to the original representation, confirming the results from Sec. \ref{sec:stax_learned_bs}.

\looseness=-1
Finally, Fig. \ref{fig:stax_split_representation} shows an example of how the learned representation for one single reward area can contain multiple distinct zones even if the ground-truth descriptor only contains one, as discussed in Sec. \ref{sec:exploitation}.
This is due to the combined effect of the representation learning done by the \gls{ae} and the stacking of multiple frames along the trajectory.
The presence of multiple reward zones in the learned representations supports the use of emitters when optimizing rewards.

\section{Discussion and Conclusion}
\label{sec:stax_conclusion}
\looseness=-1
This paper introduced \gls{name}, a method that combines the representation learning ability of \gls{taxons} \citep{paolo2020unsupervised} when dealing with unknown \gls{bs} and the capacity to focus on interesting areas of the search space of \gls{serene} \citep{paolo2021sparse} through emitters.
In addition to what \gls{taxons} does when learning \gls{bs}, \gls{name} uses multiple observations sampled along the trajectory generated by the policies to extract their behavior descriptor. 
This allows overcoming the requirement of the final observation needing to be descriptive enough to distinguish between the policies.
Moreover, by using a multi-objective approach to combine the two metrics of novelty and surprise, \gls{name} can perform better exploration compared to \gls{taxons}.
As discussed in Sec. \ref{sec:exploitation}, performing reward exploitation through emitters can prove extremely useful when exploring with a learned behavior space.
This is due to the fact that there is no guarantee that this learned space will represent all the rewards in a single connected area, as shown in Sec. \ref{sec:learned_bs}.

\looseness=-1
The results on three different sparse rewards environments show how \gls{name} can prove effective in dealing with these kinds of situations, reaching high performances both from the point of view of exploration and exploitation of the rewards.
These results are comparable to the ones obtained by \gls{serene} \citep{paolo2021sparse} notwithstanding \gls{name} being provided much less prior information about the task to solve.
Moreover, learning the behavior space while performing the search allows reducing the main limitation of \gls{ns}-based methods: the hand-design of \gls{bs}.

It is to notice that, while the choice of learning the behavior space from images might seem limiting, this is not the case.
Thanks to the simplicity and availability of cameras, many problems in robotics can be represented through images without providing problem specific information.
At the same time, images are only one type of high-dimensional observations and, while we have not tested \gls{name} on other kind of representation, there is no constraint on the type of observations to use. 
Finally, while learning the behavior space representation through an \gls{ae} introduces the need of the model design, this requires much less engineering effort than the one required to hand-design the behavior space.
This greatly increases the generalization and applicability of \gls{name}.
An example of this is the fact that to solve the three test environment we used the same \gls{ae} model structure, even if the ground truth behavior space was different.

\looseness=-1
To properly study how the aspects of policy selection and behavior space learning of \gls{name} influence the exploration process, and the discovery and exploitation of rewards, multiple ablation experiments have been performed.
The results show that the combination of using multiple observations collected during the trajectory and the multi-objective policy selection strategy are important in obtaining good coverage of the ground-truth search space.
At the same time, the continuous training of the \gls{ae} during the whole search is shown to provide a negligible curriculum effect, compared to the one provided by training on the data from the archives.
Finally, Sec. \ref{sec:learned_bs} showed how the learned space has a similar structure to the ground truth one, allowing the algorithm to perform good exploration in both.

\looseness=-1
The introduction of \gls{name} addresses the multiple shortcomings of the original \gls{ns} algorithm while at the same time opening multiple interesting avenues of research.
As for \gls{serene}, \gls{name} uses a simple scheduler to alternate between the exploration and the exploitation processes.
Applying more complex and adaptive approaches to perform the switch between the two processes can be an interesting line of work in improving the method even more.
Another possible direction of research is the one initiated by \cite{cully2020multi}, where multiple kinds of emitters are combined through a multi-armed bandit approach.
Moreover, the sampling of multiple observations along the trajectory to generate the behavior descriptor leads to interesting questions on how this sampling can be done and how the generated behaviors can be compared more meaningful ways.
Recent work started to investigate similar questions \citep{stork2020understanding, hagg2019prediction} and future work will investigate how an approach like \gls{name} can take advantage of such ideas.

\section*{Acknowledgements}
This work has received funding from the European Commission’s Horizon Europe Framework Program under grant agreement No. 101070381 (PILLAR-robots project).

\small
\bibliographystyle{apalike}
\bibliography{biblio}

\begin{thebibliography}{}

\bibitem[Andrychowicz et~al., 2017]{Andrychowicz2017HER}
Andrychowicz, M., Wolski, F., Ray, A., Schneider, J., Fong, R., Welinder, P.,
  McGrew, B., Tobin, J., Abbeel, O.~P., and Zaremba, W. (2017).
\newblock Hindsight experience replay.
\newblock In {\em Advances in Neural Information Processing Systems}, pages
  5048--5058.

\bibitem[Aubret et~al., 2019]{aubret2019survey}
Aubret, A., Matignon, L., and Hassas, S. (2019).
\newblock A survey on intrinsic motivation in reinforcement learning.
\newblock {\em arXiv preprint arXiv:1908.06976}.

\bibitem[Baranes and Oudeyer, 2013]{baranes2013active}
Baranes, A. and Oudeyer, P.-Y. (2013).
\newblock Active learning of inverse models with intrinsically motivated goal
  exploration in robots.
\newblock {\em Robotics and Autonomous Systems}, 61(1):49--73.

\bibitem[Bellemare et~al., 2016]{bellemare2016unifying}
Bellemare, M., Srinivasan, S., Ostrovski, G., Schaul, T., Saxton, D., and
  Munos, R. (2016).
\newblock Unifying count-based exploration and intrinsic motivation.
\newblock In {\em Advances in Neural Information Processing Systems},
  volume~29, pages 1471--1479.

\bibitem[Berner et~al., 2019]{berner2019dota}
Berner, C., Brockman, G., Chan, B., Cheung, V., Debiak, P., Dennison, C.,
  Farhi, D., Fischer, Q., Hashme, S., Hesse, C., et~al. (2019).
\newblock Dota 2 with large scale deep reinforcement learning.
\newblock {\em arXiv preprint arXiv:1912.06680}.

\bibitem[Burda et~al., 2018]{burda2018exploration}
Burda, Y., Edwards, H., Storkey, A., and Klimov, O. (2018).
\newblock Exploration by random network distillation.
\newblock {\em arXiv preprint arXiv:1810.12894}.

\bibitem[Cideron et~al., 2020]{cideron2020qd}
Cideron, G., Pierrot, T., Perrin, N., Beguir, K., and Sigaud, O. (2020).
\newblock Qd-rl: Efficient mixing of quality and diversity in reinforcement
  learning.
\newblock {\em arXiv preprint arXiv:2006.08505}.

\bibitem[Colas et~al., 2018]{colas2018gep}
Colas, C., Sigaud, O., and Oudeyer, P.-Y. (2018).
\newblock Gep-pg: Decoupling exploration and exploitation in deep reinforcement
  learning algorithms.
\newblock In {\em International Conference on Machine Learning}, pages
  1039--1048. PMLR.

\bibitem[Cully, 2019]{cully2019autonomous}
Cully, A. (2019).
\newblock Autonomous skill discovery with quality-diversity and unsupervised
  descriptors.
\newblock In {\em Proceedings of the Genetic and Evolutionary Computation
  Conference}, pages 81--89.

\bibitem[Cully, 2021]{cully2020multi}
Cully, A. (2021).
\newblock Multi-emitter map-elites: improving quality, diversity and data
  efficiency with heterogeneous sets of emitters.
\newblock In {\em Proceedings of the Genetic and Evolutionary Computation
  Conference}, pages 84--92.

\bibitem[Cully et~al., 2015]{CUlly2015MAPElites}
Cully, A., Clune, J., Tarapore, D., and Mouret, J.-B. (2015).
\newblock Robots that can adapt like animals.
\newblock {\em Nature}, 521(7553):503.

\bibitem[Cully and Demiris, 2017]{cully2017quality}
Cully, A. and Demiris, Y. (2017).
\newblock Quality and diversity optimization: A unifying modular framework.
\newblock {\em IEEE Transactions on Evolutionary Computation}, 22(2):245--259.

\bibitem[Deb et~al., 2002]{deb2002fast}
Deb, K., Pratap, A., Agarwal, S., and Meyarivan, T. (2002).
\newblock A fast and elitist multiobjective genetic algorithm: Nsga-ii.
\newblock {\em IEEE transactions on evolutionary computation}, 6(2):182--197.

\bibitem[Ecoffet et~al., 2019]{Ecoffet2019GO_Explore}
Ecoffet, A., Huizinga, J., Lehman, J., Stanley, K.~O., and Clune, J. (2019).
\newblock Go-explore: a new approach for hard-exploration problems.
\newblock {\em arXiv preprint arXiv:1901.10995}.

\bibitem[Eysenbach et~al., 2018]{Eysenbach2018DIAYN}
Eysenbach, B., Gupta, A., Ibarz, J., and Levine, S. (2018).
\newblock Diversity is all you need: Learning skills without a reward function.
\newblock {\em arXiv preprint arXiv:1802.06070}.

\bibitem[Fontaine et~al., 2020]{fontaine2020covariance}
Fontaine, M.~C., Togelius, J., Nikolaidis, S., and Hoover, A.~K. (2020).
\newblock Covariance matrix adaptation for the rapid illumination of behavior
  space.
\newblock In {\em Proceedings of the 2020 genetic and evolutionary computation
  conference}, pages 94--102.

\bibitem[Forestier et~al., 2022]{Forestier2017IMGEP}
Forestier, S., Portelas, R., Mollard, Y., and Oudeyer, P.-Y. (2022).
\newblock Intrinsically motivated goal exploration processes with automatic
  curriculum learning.
\newblock {\em J. Mach. Learn. Res.}

\bibitem[Gaier et~al., 2019]{gaier2019quality}
Gaier, A., Asteroth, A., and Mouret, J.-B. (2019).
\newblock Are quality diversity algorithms better at generating stepping stones
  than objective-based search?
\newblock In {\em Proceedings of the Genetic and Evolutionary Computation
  Conference Companion}, pages 115--116.

\bibitem[Grillotti and Cully, 2022]{grillotti2021unsupervised}
Grillotti, L. and Cully, A. (2022).
\newblock Unsupervised behaviour discovery with quality-diversity optimisation.
\newblock {\em IEEE Transactions on Evolutionary Computation}.

\bibitem[Hagg et~al., 2020]{hagg2020analysis}
Hagg, A., Preuss, M., Asteroth, A., and B{\"a}ck, T. (2020).
\newblock An analysis of phenotypic diversity in multi-solution optimization.
\newblock In {\em International Conference on Bioinspired Methods and Their
  Applications}, pages 43--55. Springer.

\bibitem[Hagg et~al., 2019]{hagg2019prediction}
Hagg, A., Zaefferer, M., Stork, J., and Gaier, A. (2019).
\newblock Prediction of neural network performance by phenotypic modeling.
\newblock In {\em Proceedings of the Genetic and Evolutionary Computation
  Conference Companion}, pages 1576--1582.

\bibitem[Hansen, 2016]{hansen2016cma}
Hansen, N. (2016).
\newblock The cma evolution strategy: A tutorial.
\newblock {\em arXiv preprint arXiv:1604.00772}.

\bibitem[Herdin et~al., 2005]{herdin2005correlation}
Herdin, M., Czink, N., Ozcelik, H., and Bonek, E. (2005).
\newblock Correlation matrix distance, a meaningful measure for evaluation of
  non-stationary mimo channels.
\newblock In {\em 2005 IEEE 61st Vehicular Technology Conference}, volume~1,
  pages 136--140. IEEE.

\bibitem[Holm, 1979]{holm1979simple}
Holm, S. (1979).
\newblock A simple sequentially rejective multiple test procedure.
\newblock {\em Scandinavian journal of statistics}, pages 65--70.

\bibitem[Hu et~al., 2020]{hu2020learning}
Hu, Y., Wang, W., Jia, H., Wang, Y., Chen, Y., Hao, J., Wu, F., and Fan, C.
  (2020).
\newblock Learning to utilize shaping rewards: A new approach of reward
  shaping.
\newblock {\em Advances in Neural Information Processing Systems},
  33:15931--15941.

\bibitem[Kingma and Ba, 2014]{kingma2014adam}
Kingma, D.~P. and Ba, J. (2014).
\newblock Adam: A method for stochastic optimization.
\newblock {\em arXiv preprint arXiv:1412.6980}.

\bibitem[Klambauer et~al., 2017]{Klambauer2017selu}
Klambauer, G., Unterthiner, T., Mayr, A., and Hochreiter, S. (2017).
\newblock Self-normalizing neural networks.
\newblock In {\em Advances in neural information processing systems}, pages
  971--980.

\bibitem[Laversanne-Finot et~al., 2018]{laversanne2018curiosity}
Laversanne-Finot, A., Pere, A., and Oudeyer, P.-Y. (2018).
\newblock Curiosity driven exploration of learned disentangled goal spaces.
\newblock In {\em Conference on Robot Learning}, pages 487--504. PMLR.

\bibitem[Lehman and Stanley, 2008]{lehman2008ns}
Lehman, J. and Stanley, K.~O. (2008).
\newblock Exploiting open-endedness to solve problems through the search for
  novelty.
\newblock In {\em ALIFE}, pages 329--336.

\bibitem[Lehman and Stanley, 2011]{lehman2011evolving}
Lehman, J. and Stanley, K.~O. (2011).
\newblock Evolving a diversity of virtual creatures through novelty search and
  local competition.
\newblock In {\em Proceedings of the 13th annual conference on Genetic and
  evolutionary computation}, pages 211--218. ACM.

\bibitem[Liapis et~al., 2013]{Liapis2013Delenox}
Liapis, A., Mart{\'\i}nez, H.~P., Togelius, J., and Yannakakis, G.~N. (2013).
\newblock Transforming exploratory creativity with delenox,.
\newblock In {\em ICCC}, pages 56--63.

\bibitem[Loviken and Hemion, 2017]{loviken2017online}
Loviken, P. and Hemion, N. (2017).
\newblock Online-learning and planning in high dimensions with finite element
  goal babbling.
\newblock In {\em 2017 Joint IEEE International Conference on Development and
  Learning and Epigenetic Robotics (ICDL-EpiRob)}, pages 247--254. IEEE.

\bibitem[Mann and Whitney, 1947]{mann1947test}
Mann, H.~B. and Whitney, D.~R. (1947).
\newblock On a test of whether one of two random variables is stochastically
  larger than the other.
\newblock {\em The annals of mathematical statistics}, pages 50--60.

\bibitem[Mataric, 1994]{mataric1994reward}
Mataric, M.~J. (1994).
\newblock Reward functions for accelerated learning.
\newblock In {\em Machine learning proceedings 1994}, pages 181--189. Elsevier.

\bibitem[Mouret and Clune, 2015]{mouret2015illuminating}
Mouret, J.-B. and Clune, J. (2015).
\newblock Illuminating search spaces by mapping elites.
\newblock {\em arXiv preprint arXiv:1504.04909}.

\bibitem[Nair et~al., 2018]{Nair2018ImaginedGoal}
Nair, A.~V., Pong, V., Dalal, M., Bahl, S., Lin, S., and Levine, S. (2018).
\newblock Visual reinforcement learning with imagined goals.
\newblock In {\em Advances in Neural Information Processing Systems}, pages
  9191--9200.

\bibitem[Ng et~al., 1999]{ng1999policy}
Ng, A.~Y., Harada, D., and Russell, S. (1999).
\newblock Policy invariance under reward transformations: Theory and
  application to reward shaping.
\newblock In {\em Icml}, volume~99, pages 278--287.

\bibitem[Oudeyer and Kaplan, 2009]{oudeyer2009intrinsic}
Oudeyer, P.-Y. and Kaplan, F. (2009).
\newblock What is intrinsic motivation? a typology of computational approaches.
\newblock {\em Frontiers in neurorobotics}, 1:6.

\bibitem[Paolo, 2020]{paolo2020billiard}
Paolo, G. (2020).
\newblock Billiard.
\newblock \url{https://github.com/GPaolo/Billiard}.

\bibitem[Paolo et~al., 2021]{paolo2021sparse}
Paolo, G., Coninx, A., Doncieux, S., and Laflaqui{\`e}re, A. (2021).
\newblock Sparse reward exploration via novelty search and emitters.
\newblock In {\em The Genetic and Evolutionary Computation Conference 2021
  (GECCO 2021)}.

\bibitem[Paolo et~al., 2020]{paolo2020unsupervised}
Paolo, G., Laflaquiere, A., Coninx, A., and Doncieux, S. (2020).
\newblock Unsupervised learning and exploration of reachable outcome space.
\newblock In {\em 2020 IEEE International Conference on Robotics and Automation
  (ICRA)}, pages 2379--2385. IEEE.

\bibitem[Pugh et~al., 2016]{pugh2016qdfontier}
Pugh, J.~K., Soros, L.~B., and Stanley, K.~O. (2016).
\newblock Quality diversity: A new frontier for evolutionary computation.
\newblock {\em Frontiers in Robotics and AI}, 3:40.

\bibitem[Salehi et~al., 2021]{salehi2021br}
Salehi, A., Coninx, A., and Doncieux, S. (2021).
\newblock Br-ns: an archive-less approach to novelty search.
\newblock In {\em Proceedings of the Genetic and Evolutionary Computation
  Conference}, pages 172--179.

\bibitem[Sigaud, 2022]{sigaud2022combining}
Sigaud, O. (2022).
\newblock Combining evolution and deep reinforcement learning for policy
  search: a survey.
\newblock {\em arXiv preprint arXiv:2203.14009}.

\bibitem[Stork et~al., 2020]{stork2020understanding}
Stork, J., Zaefferer, M., Bartz-Beielstein, T., and Eiben, A. (2020).
\newblock Understanding the behavior of reinforcement learning agents.
\newblock In {\em International Conference on Bioinspired Methods and Their
  Applications}, pages 148--160. Springer.

\bibitem[Sutton and Barto, 2018]{sutton2018reinforcement}
Sutton, R.~S. and Barto, A.~G. (2018).
\newblock {\em Reinforcement learning: An introduction}.
\newblock MIT press.

\bibitem[Trott et~al., 2019]{trott2019keeping}
Trott, A., Zheng, S., Xiong, C., and Socher, R. (2019).
\newblock Keeping your distance: Solving sparse reward tasks using
  self-balancing shaped rewards.
\newblock In {\em Advances in Neural Information Processing Systems}, pages
  10376--10386.

\end{thebibliography}

\end{document}